\documentclass[preprint,11pt,authoryear]{elsarticle}

\usepackage[T1]{fontenc}
\usepackage[utf8]{inputenc}
\usepackage{amsmath,amssymb}
\usepackage{booktabs,array}
\usepackage{graphicx}
\usepackage{xcolor}
\usepackage{tikz}
\usetikzlibrary{positioning,arrows.meta}
\usepackage[hidelinks]{hyperref}

\graphicspath{{figures/}}
\newcommand{\nRecords}{11{,}382}
\newcommand{\nTasks}{48}

\newcommand{\nPermutations}{10{,}000}

\newcommand{\Ktemp}{0.7}
\newcommand{\Krep}{5}
\newcommand{\Lpar}{2}

\newcommand{\dialGpt}{0.688}
\newcommand{\freshGpt}{0.896}
\newcommand{\diffGpt}{0.208}
\newcommand{\ciGpt}{[0.063, 0.354]}
\newcommand{\pHolmGpt}{0.026}

\newcommand{\diffLlama}{0.104}
\newcommand{\ciLlama}{[0.000, 0.208]}
\newcommand{\pHolmLlama}{0.125}

\newcommand{\dialMistral}{0.063}
\newcommand{\freshMistral}{0.354}
\newcommand{\diffMistral}{0.292}
\newcommand{\ciMistral}{[0.146, 0.438]}
\newcommand{\pHolmMistral}{0.004}

\newcommand{\nFreshOnly}{34}
\newcommand{\nDialOnly}{5}
\newcommand{\crlAuroc}{0.627}
\newcommand{\crlCi}{[0.493, 0.760]}
\newcommand{\crlP}{0.022}

\newcommand{\robAllAuroc}{0.571}
\newcommand{\robAllP}{0.057}

\newcommand{\cdrSzeroEDialRange}{0.10--0.14}
\newcommand{\cdrSzeroECtrlRange}{0.82--0.86}

\newcommand{\cdrLlamaControlGap}{0.545}

\newcommand{\cdrLlamaControlHolmP}{$<$0.001}
\newcommand{\cdrLlamaControlLateGap}{0.575}
\newcommand{\cdrGemmaControlGap}{0.588}

\newcommand{\cdrGemmaControlHolmP}{$<$0.001}
\newcommand{\cdrGemmaControlLateGap}{0.630}
\newcommand{\cdrLevelMatched}{-0.118}
\newcommand{\cdrLevelMatchedP}{$<$0.001}
\newcommand{\cdrLevelMatchedN}{90}
\newcommand{\cdrLevelMatchedSzeroE}{-0.185}

\newcommand{\valLlamaDetect}{6/18}

\newcommand{\valLlamaLsSignal}{11/18}
\newcommand{\valLlamaLsNull}{6/12}

\newcommand{\valLlamaAggNull}{0/2}
\newcommand{\valGemmaDetect}{1/18}

\newcommand{\valGemmaLsSignal}{16/18}
\newcommand{\valGemmaLsNull}{3/12}

\newcommand{\valGemmaAggNull}{1/2}
\newcommand{\replaySsDeltaGpt}{+0.46}
\newcommand{\replaySsDeltaLlama}{+0.42}
\newcommand{\replayLsDeltaGpt}{+0.31}
\newcommand{\silentCQEarly}{0.61}
\newcommand{\silentCQLate}{0.08}
\newcommand{\silentCSEarly}{0.79}
\newcommand{\silentCSLate}{0.64}
\newcommand{\silentLSEarly}{0.54}
\newcommand{\silentLSLate}{0.47}
\newcommand{\silentSSEarly}{0.94}
\newcommand{\silentSSLate}{0.98}
\newcommand{\silentCsSupport}{2976}

\newcommand{\crlLastGapAuroc}{0.672}
\newcommand{\crlLastGapP}{0.001}
\newcommand{\crlCondAuroc}{0.468}
\newcommand{\crlCondP}{0.674}



\newdefinition{definition}{Definition}
\newtheorem{proposition}{Proposition}
\newproof{pf}{Proof}

\journal{Information Sciences}

\begin{document}

\begin{frontmatter}

\title{The Silent Cost of Conversation History in LLM Reasoning}

\author[tarsus]{Ali \c{S}enol\corref{cor1}}
\ead{alisenol@tarsus.edu.tr}
\cortext[cor1]{Corresponding author.}
\affiliation[tarsus]{organization={Department of Computer Engineering, Tarsus University},
  city={Tarsus, Mersin}, postcode={33400}, country={T\"urkiye}}

\begin{abstract}
Large language models (LLMs) in multi-turn information systems accumulate
their own earlier replies and a growing task state. We asked whether this
conversation history lowers accuracy, and whether process-level quality
metrics warn before it does. Each analysis was fixed before its outcomes were
computed. In Study~1, on 48 sharded math tasks, retaining its replies
lowered final success by 0.21 for GPT-4o-mini and 0.29 for Mistral-7B
(Holm-adjusted $p=0.026$ and $0.004$); Llama-4 Scout showed the same
direction, not significant. Study~2 introduces CDR-Stream, 100-turn
state-tracking dialogues with an exact answer per turn and per-model
difficulty calibration. With the full history, accuracy fell from about 0.6 to
0.1 within twenty turns at matched difficulty, whereas a solver-verified state
summary kept it 0.52--0.59 higher (Holm-adjusted $p<0.001$). Consistency and
logical coherence fell far less than accuracy, and stability rose. None
of these metrics, a paired readiness measure or calibrated single-stream drift
detectors gave robust, model-general early warning, although in Study~2 the
repeat-based metrics were complete only for GPT-4o-mini. In the monitors we
tested, a single 100-turn stream had power only for very large drops. Systems
built on LLMs should keep task state outside the transcript and monitor across
sessions.
\end{abstract}

\begin{keyword}
large language models \sep multi-turn dialogue \sep conversation history \sep
state tracking \sep reasoning quality \sep drift detection \sep early warning
\end{keyword}

\end{frontmatter}

\section{Introduction}
\label{sec:introduction}

LLMs are increasingly used as components of information and decision-support
systems that keep a task open over many turns. A procurement assistant tracks a
budget as invoices arrive; a logistics assistant follows stock across
warehouses. The user supplies information piece by piece and asks questions
whose answers depend on everything said so far, so the model's context fills
with its own earlier replies and with a task state that changes at every turn.
We call this accumulated context the \emph{conversation history}, and the
experimental condition that keeps it \emph{dialogue history}.

\paragraph{Why this is a reliability problem, not only an accuracy problem}
A growing conversation history can let a model lock onto a wrong task state:
once an early turn is answered incorrectly, every later reply can be generated
consistently from that wrong state, without the model producing any of the
usual symptoms of failure. If a deployed system does not see the gold answer,
and the responses it does see remain fluent, mutually consistent and locally
coherent, the failure is invisible unless some other signal flags it. The paper
asks two questions that follow from this chain: does retaining history cost
accuracy, and does that cost compound as the conversation grows; and if it
does, can the cost be seen before it shows up in the final answer, using
signals a deployed system can actually compute?

\paragraph{What is known}
Multi-turn use is known to be costly. Models lose accuracy when a task is
revealed over several turns because they commit to premature answers
\citep{laban2025lost}; earlier assistant replies can propagate errors
\citep{huang2026benefit}; and the effective context is shorter than the
nominal one \citep{liu2024lost,hsieh2024ruler}.

\paragraph{Two gaps these results leave open}
First, a \emph{measurement gap}: sharded-task studies such as \citet{laban2025lost}
score a single answer after at most a handful of turns, so they show that
history can hurt but not how the loss develops over a long horizon. Over tens of turns, the questions themselves
may also become harder, so a falling accuracy curve can reflect the questions
rather than the context unless difficulty is explicitly held fixed. Second, a
\emph{monitoring gap}: none of these studies treats reasoning-quality metrics
as a stream, applies a detector calibrated on a stationary baseline, or reports
a false-alarm rate. Showing that performance eventually drops is not the same
as showing that the drop can be seen coming. ERGO is the closest exception, but
it monitors token-entropy spikes rather than response-level quality metrics,
and it is not evaluated against a long, difficulty-controlled horizon
\citep{khalid2025ergo}. To our knowledge, no prior study jointly (i) holds
difficulty fixed over a long multi-turn horizon and (ii) tests established
quality metrics as calibrated early-warning signals with a controlled
false-alarm rate.

\paragraph{This paper}
We close both gaps with two experiments, each analysed under a protocol fixed
before its outcomes were computed. Study~1 addresses the short-horizon
measurement gap directly: it compares, turn by turn, the same user information
with and without the model's own earlier replies. Study~2 addresses the
long-horizon and monitoring gaps together with CDR-Stream, a benchmark of
100-turn state-tracking dialogues with an exact answer at every turn,
difficulty calibrated per model, and a control arm that replaces the history
with a solver-verified state summary. The two studies differ in tasks, models
and controls, so we read them as two related failure modes rather than as one
dose--response curve.

The paper makes three contributions:
\begin{enumerate}
  \item a prospectively frozen, paired estimate of the cost of retaining
    assistant replies on short sharded tasks, which replicates the direction
    reported by \citet{huang2026benefit};
  \item CDR-Stream, a long-horizon, difficulty-calibrated benchmark in which
    accuracy under full history collapsed within about twenty turns at matched
    difficulty, while a verified state summary preserved it;
  \item a prospectively defined evaluation of whether established
    reasoning-quality metrics and calibrated single-stream detectors can
    serve as early warnings of this collapse -- assessing not only whether
    they move before accuracy fails, but whether any such lead survives a
    controlled false-alarm rate and repeats across models. None did: coherence
    and stability moved the wrong way, staying flat or rising as accuracy
    fell, and a closed-form bound together with a power simulation show that,
    for the detectors we tested, much of this reflects what a single 100-turn
    conversation can reveal rather than a property of the metrics themselves.
\end{enumerate}

\section{Related work}
\label{sec:related}

\paragraph{Multi-turn and long-context degradation}
Benchmarks such as MT-Bench, MT-Eval and MINT evaluate conversational ability
over several exchanges \citep{zheng2023judging,kwan2024mteval,wang2024mint}.
\citet{laban2025lost} sharded fully specified tasks across turns and traced
large losses to premature answers on which models then anchored.
\citet{huang2026benefit} omitted or condensed assistant turns, including on the
Lost in Conversation math tasks, and found that earlier replies can propagate
errors, and \citet{sinha2025illusion} reported self-conditioning on earlier
errors in long-horizon execution. Related mechanisms include information lost
in the middle of long inputs \citep{liu2024lost}, effective context shorter
than advertised \citep{hsieh2024ruler}, distraction by irrelevant context
\citep{shi2023distracted}, sycophancy \citep{sharma2024sycophancy}, unreliable
self-correction \citep{huang2024cannot} and difficulty tracking entity states
\citep{kim2023entity}. On the mitigation side, memory architectures summarise
or page earlier content and are tested on whether facts survive across
sessions \citep{packer2023memgpt,zhong2024memorybank,wang2023recursive,
maharana2024locomo,wu2025longmemeval}, and ERGO rebuilds the prompt after
spikes in token entropy \citep{khalid2025ergo}; our verified state summary is
an upper reference for such compact memory, not a deployable summariser.

This work documents three distinct reasons accuracy might fall -- lost
mid-input information, inconsistent task-state tracking, and anchoring on
one's own earlier replies -- but not whether any of the three can be seen,
without the gold answer, before it shows up in the final answer; none of the
cited studies calibrates a detector against such a signal with a controlled
false-alarm rate.

\paragraph{Reasoning quality and drift monitoring}
Reasoning quality has been characterised by agreement among sampled solutions
\citep{wang2023selfconsistency}, faithfulness of chains of thought
\citep{wei2022chain,turpin2023say,lanham2023measuring} and models' knowledge of
their own correctness \citep{kadavath2022know}; sample checks
\citep{manakul2023selfcheckgpt} and semantic entropy
\citep{farquhar2024detecting} estimate uncertainty without labels. Recent work
adapts clinical cognitive tests to LLMs \citep{park2027lmmse} and optimises the
steps of mathematical reasoning \citep{lee2026fgrpo}. These approaches judge one
response at a time; we instead ask whether the same signals, aggregated per
turn, anticipate a later failure. We use the six RQEval dimensions
\citep{rqeval2026} unchanged, rather than proposing new ones, because three of
them are already label-free candidate precursors: CS and SS compare repeated
samples of the same reply, and LS scores coherence within one reasoning trace,
none needing the gold answer (unlike CQ and ES, the outcome itself, and RS,
which needs the gold to judge a reworded reply against a \emph{correct}
primary answer). CUSUM
\citep{page1954continuous}, Page--Hinkley
\citep{hinkley1971inference}, EWMA \citep{roberts1959control}, DDM
\citep{gama2004learning} and ADWIN \citep{bifet2007learning} are standard
stream detectors whose delay and false alarms have been compared extensively
\citep{gama2014survey,webb2016characterizing,goncalves2014comparative,barros2018largescale,bayram2022concept};
recent methods classify drift types \citep{deng2025zilean} and couple
detection with adaptation \citep{abushaira2027sccm}. Such detectors usually
watch input distributions or labelled performance
\citep{sculley2015hidden,rabanser2019failing}, and commercial LLM behaviour is
known to drift \citep{chen2024chatgpt}. We apply them instead to per-turn
reasoning-quality streams within a conversation.

Table~\ref{tab:positioning} places our studies against the closest work. We do
not claim to be the first to show that assistant history can hurt; we add a
prospectively frozen paired estimate, exact per-turn answers over 100 turns at
calibrated difficulty, and a direct test of quality metrics and detectors as
early warnings.

\begin{table}[!htbp]
\centering
\caption{Positioning against the closest prior work.}
\label{tab:positioning}
\footnotesize
\setlength{\tabcolsep}{4pt}
\begin{tabular}{@{}>{\raggedright\arraybackslash}p{0.19\linewidth}
>{\raggedright\arraybackslash}p{0.25\linewidth}
>{\raggedright\arraybackslash}p{0.25\linewidth}
>{\raggedright\arraybackslash}p{0.25\linewidth}@{}}
\toprule
Study & Setting & Role of history & Signals for monitoring\\
\midrule
\citet{laban2025lost} & sharded single-turn tasks & cause of premature,
unreliable answers & not studied\\
\citet{huang2026benefit} & real and sharded conversations & assistant turns
omitted or condensed & not studied\\
\citet{sinha2025illusion} & synthetic long-horizon execution & self-conditioning
on own errors & not studied\\
ERGO \citep{khalid2025ergo} & sharded tasks & reset after uncertainty spikes &
token entropy as a trigger\\
Study 1 (ours) & sharded math tasks, paired arms & net effect of retaining
replies & CS, LS, SS and a paired readiness measure\\
Study 2 (ours) & 100-turn state tracking, calibrated difficulty & full history
versus verified state & six RQEval metrics, five detectors\\
\bottomrule
\end{tabular}
\end{table}

\section{Monitoring framework and evaluation design}
\label{sec:framework}

Both studies share the context policies, the per-turn quality signals, the
streaming monitor and the evidence standards defined here
(Figure~\ref{fig:framework}); Table~\ref{tab:datasets} summarises the two
benchmarks before the details below. Full model identifiers, gateway aliases
and acquisition dates are in supplementary Table~S18.

\begin{table}[!htbp]
\centering
\caption{The two benchmarks at a glance. K/L: same-context repeats / reworded
instructions per turn. Metric coverage lists which of the six dimensions were
computed where; CQ, RS and ES always need a final answer and the gold
(Section~\ref{sec:framework}).}
\label{tab:datasets}
\footnotesize
\setlength{\tabcolsep}{4pt}
\begin{tabular}{@{}>{\raggedright\arraybackslash}p{0.17\linewidth}
>{\raggedright\arraybackslash}p{0.40\linewidth}
>{\raggedright\arraybackslash}p{0.40\linewidth}@{}}
\toprule
 & Study 1 (Lost in Conversation) & Study 2 (CDR-Stream)\\
\midrule
Source & GSM8K math, sharded \citep{cobbe2021training,laban2025lost} &
purpose-built, solver-checked\\
Units & \nTasks{} tasks, 4--9 turns (median 6) & 30 dialogues/model, 100 turns
each\\
Domains & arithmetic word problems & budget, inventory, lab, milestones (4)\\
Difficulty & not controlled & 12-level ladder, calibrated per model\\
Scenarios & single ordered reveal & 5 (2 stationary null, 3 scheduled-change)\\
Models & GPT-4o-mini, Llama-4 Scout, Mistral-7B & GPT-4o-mini, Llama-3 70B,
Gemma 3 12B\\
$K$ / $L$ & \Krep{} / \Lpar{}, all models & 3 / 2, GPT-4o-mini only\\
Metric coverage & CQ/RS/ES at the final turn; CS/LS/SS at pre-final turns
(gold not yet available) & CQ/LS/ES every turn, all three models; CS/RS/SS
every turn, GPT-4o-mini only (repeat-based)\\
\bottomrule
\end{tabular}
\end{table}

\begin{figure}[!htbp]
\centering
\begin{tikzpicture}[
  font=\small, >=Latex, node distance=4mm and 6mm,
  box/.style={draw, rounded corners=2pt, align=center, minimum height=9mm,
              inner sep=3pt, fill=#1},
  box/.default=white]
\node[box=gray!10, text width=27mm] (user) {User turn $u_t$\\(new information,\\question)};
\node[box=blue!8, text width=33mm, right=of user] (policy)
  {Context policy $\pi$\\[1pt]\footnotesize dialogue history $\cdot$ fresh prefix\\
   \footnotesize $\cdot$ verified state summary};
\node[box=gray!10, text width=24mm, right=of policy] (llm) {LLM\\primary reply\\+ $K{-}1$ samples\\+ $L$ rewordings};
\node[box=orange!10, text width=26mm, below=9mm of llm] (metrics)
  {Per-turn quality\\vector $\mathbf{q}_t$\\\footnotesize CQ CS RS LS ES SS};
\node[box=orange!10, text width=27mm, left=of metrics] (window)
  {Window means\\$\bar q_{m,t}$ over the last\\$W$ valid values};
\node[box=red!8, text width=30mm, left=of window] (detect)
  {Drift detectors\\\footnotesize CUSUM, PH, EWMA,\\\footnotesize DDM, ADWIN};
\node[box=green!10, text width=30mm, below=8mm of detect] (alarm)
  {Alarm time $\tau_m$\\\footnotesize compared with the change $c$};
\node[box=green!10, text width=27mm, right=of alarm] (null)
  {Threshold from\\stationary nulls\\\footnotesize false alarms $\le\alpha$};
\node[box=gray!5, text width=26mm, right=of null] (gold)
  {Exact gold $y_t$\\\footnotesize(evaluation only)};
\draw[->] (user) -- (policy);
\draw[->] (policy) -- (llm);
\draw[->] (llm) -- (metrics);
\draw[->] (metrics) -- (window);
\draw[->] (window) -- (detect);
\draw[->] (detect) -- (alarm);
\draw[->, dashed] (null) -- (alarm);
\draw[->, dashed] (gold) -- (null);
\draw[->, dashed] (gold.north) -- (metrics.south);
\draw[->, dotted] (llm.north) -- ++(0,4mm) -| node[pos=0.25, above, font=\footnotesize]
  {primary reply enters later history (dialogue history only)} (policy.north);
\end{tikzpicture}
\caption{The monitoring framework. A context policy decides what the model
sees at turn $t$. Responses are scored into a per-turn quality vector, smoothed
over a window and passed to drift detectors whose thresholds are calibrated on
stationary dialogues. Exact gold is needed for CQ, RS and ES and for
evaluating alarms; CS, LS and SS do not use it.}
\label{fig:framework}
\end{figure}

\subsection{Context policies}

At turn $t$ the user sends $u_t$ and the model returns a primary reply $r_t$
from a context $C_t$ set by a policy $\pi$:
\begin{align}
  \text{dialogue history:}\quad & C_t = (u_1, r_1, \dots, r_{t-1}, u_t),\\
  \text{fresh prefix:}\quad & C_t = (u_1, \dots, u_t),\\
  \text{verified state summary:}\quad & C_t = (s_{t-1}, u_t),
\end{align}
where $s_{t-1}$ is the exact task state after turn $t-1$, rendered by a
solver. Dialogue history versus fresh prefix estimates the \emph{net} effect of
keeping assistant replies; context length and message positions also differ,
so the effect is not attributed to the content of the replies alone. The
verified state summary shortens the context, removes earlier replies and
supplies a correct state at once, so it tests a mitigation, not a mechanism.
Every context is packed into the request with server-side memory, retrieval and
prompt rewriting disabled, which makes it reproducible from the stored records.

\paragraph{A worked example (CDR-Stream)}
Suppose turn 7 reports ``warehouse~B received 40 crates of part~X,'' and turn
8 asks how many crates of part~X warehouse~B now holds. Under \emph{dialogue
history}, $C_8$ is the full transcript of turns 1--8, including whatever the
model itself answered at every earlier turn, right or wrong and unflagged
either way. Under the \emph{verified state summary}, $C_8$ instead has only
the solver's current ledger line for warehouse~B/part~X plus turn 8; the model
never sees turns 1--7 or its own earlier replies. The gold answer comes from
the same solver and ledger, independently of what either arm is shown
(full generator and prompt in Section~\ref{sec:study2}).

\subsection{Per-turn quality signals}

Each primary reply comes with $K-1$ further samples from the same context and
$L$ replies to frozen rewordings of the instruction. RQEval \citep{rqeval2026}
computes six dimensions in $[0,1]$, used unchanged: accuracy (CQ, the primary
answer equals the gold), consistency (CS, pairwise agreement of the $K$
answers), robustness (RS, agreement of reworded replies with a correct primary
answer), logical coherence (LS, absence of NLI contradictions between
consecutive steps; \citealp{he2021deberta}), efficiency (ES, correctness
combined with token economy) and stability (SS, BERTScore similarity of the $K$
traces; \citealp{zhang2020bertscore}). CQ, RS and ES need a final answer and the
gold, CS needs at least two answers, and LS and SS need only reasoning text.
Only LS and SS can therefore serve as label-free warnings before a model is
ready to answer. LS splits a reply's reasoning into consecutive steps and
reports one minus the mean NLI-scored contradiction probability between
adjacent pairs, so a reply that never contradicts itself scores 1 regardless
of whether its conclusion is right; SS instead BERTScores the $K$ sampled
\emph{replies} against each other, so near-duplicate phrasing raises SS even
when the shared answer is wrong. Both need parseable reasoning text; missing
values are never imputed and are reported with their support (supplementary
Section~S1.2 gives the exact splitting and scoring rules).

\subsection{The streaming monitor}

Let $q_{m,t}\in[0,1]$ be the value of metric $m$ at turn $t$, or missing.

\begin{definition}[Window mean]
Let $V_{m,t}$ be the set of the last $W$ turns $j\le t$ at which $q_{m,j}$ is
observed. Once $|V_{m,t}|=W$, the monitored statistic is
\begin{equation}
  \bar q_{m,t} = \frac{1}{W}\sum_{j\in V_{m,t}} q_{m,j},
  \qquad W\in\{10,20,30\},
  \label{eq:windowmean}
\end{equation}
with $W=20$ primary. Missing values neither enter nor advance the window.
\end{definition}

Five detectors, CUSUM, Page--Hinkley, EWMA, DDM and ADWIN, are assigned to
metrics by three strategies fixed in the funded project design: CUSUM on every
metric, a distribution-specific mapping, and ADWIN on every metric
(supplementary Table~S8).

\begin{definition}[Alarm, calibration and detection]
For a detector with threshold $h$ reading $\bar q_{m,1},\dots,\bar q_{m,t}$,
the alarm time in a stream of $T$ turns is
\begin{equation}
  \tau_m(h) = \min\{t\le T : \text{the detector alarms at } t\},
  \label{eq:alarm}
\end{equation}
with $\tau_m=\infty$ if no alarm occurs. The threshold $h^\ast$ is the most
sensitive grid value whose first-alarm rate on stationary streams,
$\widehat{P}_0(\tau_m(h)\le T)$, is at most $\alpha$; drift data are never used.
For a change at turn $c$, an alarm is a \emph{false alarm} if
$\tau_m(h^\ast)<c$ and a \emph{detection} if $c\le\tau_m(h^\ast)\le c+33$.
\end{definition}

Study~1 estimates $\widehat{P}_0$ from permutation nulls with family-wise
$\alpha=0.05$, Study~2 from its stationary dialogues. Detecting a scheduled
change within 33 turns is not yet a warning. We call a signal an \emph{early
warning} only if, in addition, it alarms before the accuracy (CQ) alarm in the
same dialogue, keeps the null calibration, and does both across models.

\subsection{A paired readiness measure}

Sharded tasks allow an answer-free paired signal. Let $a_t^{\mathrm{F}}$ and
$a_t^{\mathrm{D}}$ be 1 if the fresh-prefix and dialogue-history arms attempt an
answer at turn $t$, and 0 if they ask for more information. The
\emph{Cumulative Resolution Lag} of a task is
\begin{equation}
  \mathrm{CRL} = \frac{1}{|\mathcal{T}|}\sum_{t\in\mathcal{T}}
  \max\!\left(a_t^{\mathrm{F}}-a_t^{\mathrm{D}},\,0\right),
\end{equation}
over the pre-final turns $\mathcal{T}$ at which both arms responded validly. It
measures how persistently the history arm lags an information-matched
counterfactual in becoming ready to answer, and it was fixed as an auxiliary
measure before the Study~1 data were collected.

\subsection{Evidence standards}

Each result carries one of four labels: \emph{confirmatory} (a hypothesis fixed
before its outcomes were computed), \emph{prespecified secondary},
\emph{exploratory} (rules fixed before computation on data already seen) or
\emph{post hoc}. In Study~1 the whole protocol, including the analysis, was
frozen before data collection. In Study~2 the design, sample, difficulty
mappings, prompt and truncation policy were frozen before the main run; the
analysis protocol was frozen when a budget pause had left 25,039 of 54,000
calls complete, before any main-run metric was computed (dated timeline in
supplementary Table~S19). Raw responses are stored before
scoring and never overwritten, failed gates are reported rather than
relabelled, and every number, table and figure is regenerated by scripts from
hashed inputs.

\section{Study 1: the paired effect of retaining assistant replies}
\label{sec:study1}

\subsection{Design}

We used the math subset of Lost in Conversation \citep{laban2025lost}: 103
GSM8K-derived problems \citep{cobbe2021training}, each split into information
shards revealed one per turn. Before any model call, an input audit excluded 17
tasks that duplicated our earlier inputs or appeared to change meaning when
sharded, and a hash-based draw that used no model output selected \nTasks{}
tasks with 4--9 shards (median 6). GPT-4o-mini \citep{openai2024gpt4o}, Llama-4
Scout and Mistral-7B \citep{jiang2023mistral} answered every turn in both the
dialogue-history and fresh-prefix arms. Each reply had to end with
\texttt{FINAL\_ANSWER: <number>} or \texttt{STATUS: NEED\_MORE\_INFORMATION}.
Each cell received $K=\Krep{}$ samples and $L=\Lpar{}$ rewordings at
temperature \Ktemp{}; only the primary reply entered later history. All
\nRecords{} calls succeeded without truncation; six empty Mistral-7B primaries
were kept as invalid.

The primary endpoint was paired final-turn success, tested by exact McNemar
tests \citep{mcnemar1947note} with task-bootstrap 95\% intervals
(\nPermutations{} resamples; \citealp{efron1979bootstrap}) and Holm correction
across models \citep{holm1979simple}. Before the last shard most tasks have no
determined answer, so CQ, RS and ES cannot be observed; the pre-final warning
candidates were CS, LS and SS. Each was tested by a task-clustered partial
correlation with final failure, adjusted for model, turn, revealed information
and log context length, with Holm correction across the three metrics. CRL
(defined in Section~\ref{sec:framework}) measures delay in becoming ready to
answer, not accuracy; it targeted a positive final readiness gap, in which the
fresh prefix answers while the history arm is still waiting for information.

\subsection{Results}

\begin{figure}[!htbp]
\centering
\includegraphics[width=\linewidth]{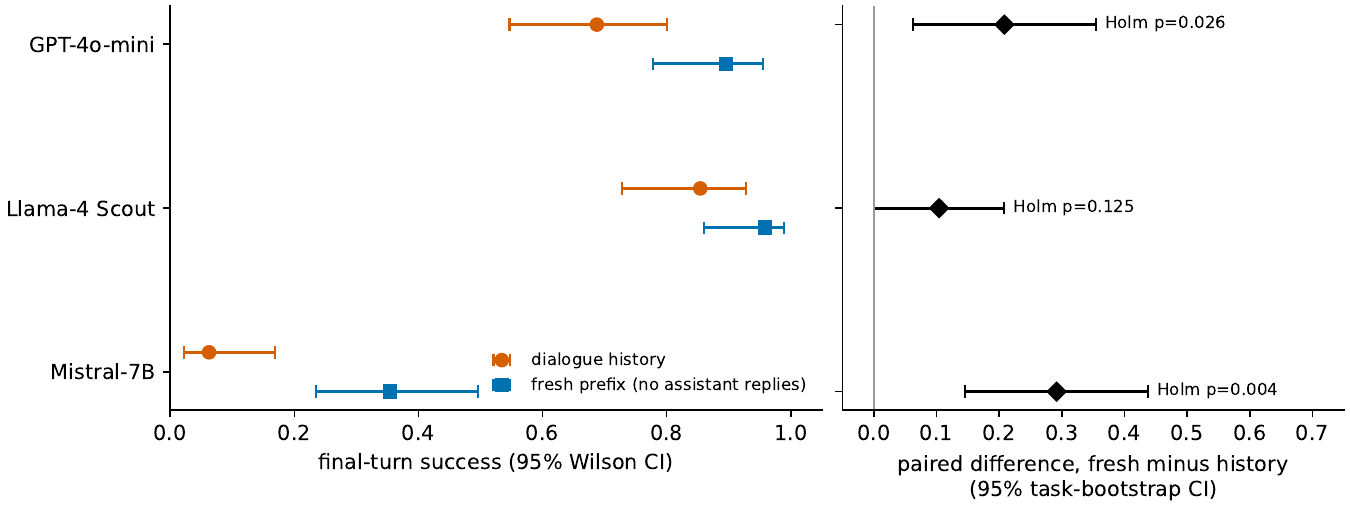}
\caption{Study 1 primary endpoint ($n=\nTasks{}$ paired tasks per model).
Left: final-turn success with 95\% Wilson intervals \citep{wilson1927probable}.
Right: paired difference, fresh prefix minus dialogue history, with 95\%
task-bootstrap intervals and Holm-adjusted McNemar $p$-values.}
\label{fig:s1}
\end{figure}

\paragraph{Retaining assistant replies lowers final success (confirmatory)}
Final success fell from \freshGpt{} to \dialGpt{} for GPT-4o-mini (difference
\diffGpt{}, 95\% CI \ciGpt{}, Holm $p=\pHolmGpt$) and from \freshMistral{} to
\dialMistral{} for Mistral-7B (\diffMistral{}, CI \ciMistral{}, Holm
$p=\pHolmMistral$). Llama-4 Scout moved in the same direction (\diffLlama{}, CI
\ciLlama{}, Holm $p=\pHolmLlama$) without significance
(Table~\ref{tab:s1primary}, Figure~\ref{fig:s1}). Across models, only the fresh
prefix succeeded in \nFreshOnly{} model--task pairs and only the history arm in
\nDialOnly{}. The GPT-4o-mini and Mistral-7B effects remained when tasks with
format-invalid final replies were excluded (supplementary
Table~S1). For Mistral-7B the harm took a visible form: with
its own replies in context it tended to stay in the ``need more information''
state at the last turn, whereas without them it answered.

\begin{table}[!htbp]
\centering
\caption{Study 1 paired final-turn success over \nTasks{} tasks per model.
F/D: tasks in which only the fresh prefix, or only the dialogue history,
succeeded; $p$: exact McNemar test.}
\label{tab:s1primary}
\small
\resizebox{\linewidth}{!}{%
\begin{tabular}{lcccccc}
\toprule
Model & Dialogue history & Fresh prefix & Diff.\ [95\% CI] & F/D & $p$ & $p_{\text{Holm}}$\\
\midrule
GPT-4o-mini & 33/48 (0.688) & 43/48 (0.896) & 0.208 [0.063, 0.354] & 12/2 & 0.013 & 0.026 \\
Llama-4 Scout & 41/48 (0.854) & 46/48 (0.958) & 0.104 [0.000, 0.208] & 6/1 & 0.125 & 0.125 \\
Mistral-7B & 3/48 (0.063) & 17/48 (0.354) & 0.292 [0.146, 0.438] & 16/2 & 0.001 & 0.004 \\
 
\bottomrule
\end{tabular}
}
\end{table}

\paragraph{The measurable quality metrics do not warn (confirmatory)}
CS, LS and SS showed no statistically detectable association with final
failure beyond turn, revealed information and context length: all partial
correlations were close to zero and every Holm-adjusted $p$-value was 1.0
(Table~\ref{tab:s1rqeval}). This does not mean that they carry no information;
CS was measurable on only 29 source tasks, so small associations cannot be
excluded.

\begin{table}[!htbp]
\centering
\caption{Study 1 pre-final metrics as predictors of final failure under
dialogue history. A unit is one (task, model) pair (up to $3\times\nTasks{}$);
turns are the pre-final turns of those units pooled together, the level at
which the metric is observed; tasks are the distinct source tasks contributing
at least one turn, the level at which the permutation test clusters
observations. $r_{\text{partial}}$: task-clustered partial correlation; $p$:
one-sided task-clustered permutation test.}
\label{tab:s1rqeval}
\small
\begin{tabular}{lcccccc}
\toprule
Metric & Turns & Units & Tasks & $r_{\text{partial}}$ & $p$ & $p_{\text{Holm}}$\\
\midrule
CS & 101 & 60 & 29 & -0.047 & 0.571 & 1.000 \\
LS & 637 & 144 & 48 & +0.017 & 0.365 & 1.000 \\
SS & 627 & 144 & 48 & -0.021 & 0.653 & 1.000 \\
 
\bottomrule
\end{tabular}
\end{table}

\paragraph{The paired readiness measure reduces to persistence}
CRL discriminated its target modestly (pooled model-stratified AUROC
\crlAuroc{}, permutation $p=\crlP$, CI \crlCi{}; supplementary
Table~S2). Post hoc, the last pre-final readiness gap alone did
better (AUROC \crlLastGapAuroc{}, $p=\crlLastGapP$), and within its strata CRL
added nothing (AUROC \crlCondAuroc{}, $p=\crlCondP$). Against accuracy loss, a
post-hoc target, the association was borderline (AUROC \robAllAuroc{},
$p=\robAllP$) and came from Mistral-7B alone. Even a paired, counterfactual
measure thus recorded the failure once the history arm had fallen behind
rather than anticipating it.

\section{Study 2: long-horizon degradation under calibrated difficulty}
\label{sec:study2}

Study~1 tasks span at most nine turns, and their difficulty is not controlled.
In long dialogues the questions themselves may become harder, so a falling
accuracy curve could reflect the questions rather than the context. Study~2
therefore uses a purpose-built benchmark with an exact answer at every turn and
difficulty set by design.

\subsection{The CDR-Stream benchmark and design}

\paragraph{Dialogues}
CDR-Stream contains 100-turn dialogues in four state-tracking domains: a budget
ledger, inventory across warehouses, a laboratory protocol with container
volumes, and chains of project milestones. Each user turn states events that
update a shared state, such as a delivery, a transfer or a correction, possibly
with distractor sentences, and asks one question: a sum, a difference, a count
above a threshold, a maximum, a what-if or a share. An exact solver computes
the gold answer of every turn.

\paragraph{Calibrated difficulty and scenarios}
Difficulty follows a 12-level ladder that changes one axis per step: query
terms, events per turn, distractors or event complexity (supplementary
Table~S4). Before any stream existed, a single-turn calibration
mapped easy, medium and hard regimes to ladder levels for each model and domain,
targeting accuracy bands of 0.75--0.92, 0.60--0.78 and 0.35--0.65, and a frozen
rule (isotonic fit, \citealp{barlow1972statistical}; one-sided Cochran--Armitage
trend test, \citealp{armitage1955tests}; minimum regime gaps) excluded domains
with a ceiling, a floor or no trend (supplementary Table~S5). Each
model is thus evaluated in its own informative range, at the cost of models not
all being tested on the same set of domains; CDR-Stream is not designed to
rank models against each other, and we do not compare raw accuracy across
models for this reason. Five scenarios fix the
regime sequence: stationary easy (S0-E) and hard (S0-H), which serve as
monitoring nulls; a staircase from easy to hard with changes at turns 34 and 68
(S1); a sudden easy-to-hard shift at turn 51 (S2); and easy to hard to easy
with changes at 34 and 67 (S3); within a regime, levels are mixed around the
centre level (supplementary Section~S2.2).

\paragraph{Models, arms and analysis}
GPT-4o-mini, Llama-3 70B \citep{grattafiori2024llama3} and Gemma 3 12B
\citep{gemma2025gemma3} were retained because they passed calibration and a
technical pilot, before any stream outcome existed; replaced candidates and two
disclosed deviations for Llama-3 70B are documented in the supplement. Each
model received 30 held-out dialogues, six per scenario, in two arms: dialogue
history, and the verified state summary with the current turn. Both arms used the same frozen prompt and response contract (verbatim in
supplementary Section~S2.1), temperature 0.7 and a 2,048-token
ceiling.
The dialogue arm also drew $K=3$ samples and $L=2$ rewordings per turn;
because the research budget staged acquisition, these were completed only for
GPT-4o-mini. Truncated replies were kept and scored as missing under a policy
frozen before the main run. Dialogues are the independent units. A
prespecified secondary contrast compares accuracy at turns 1--50 and 51--100 at
the same ladder level within each dialogue -- holding question-side difficulty
fixed (query terms, events per turn, distractors and event complexity), not
the conversation length or the accumulated state, since those grow across
turns under both arms by design. The confirmatory hypothesis H-C1 states that,
for Llama-3 70B and Gemma 3 12B, per-dialogue accuracy over turns 1--100 is
higher under the verified state summary, tested by one-sided sign-flip tests
over 30 dialogues (20,000 draws;
\citealp{good2005permutation}) with bootstrap intervals and Holm correction.
GPT-4o-mini control data had been seen before this protocol and are reported
outside the family.

\subsection{Results}

\begin{figure}[!htbp]
\centering
\includegraphics[width=\linewidth]{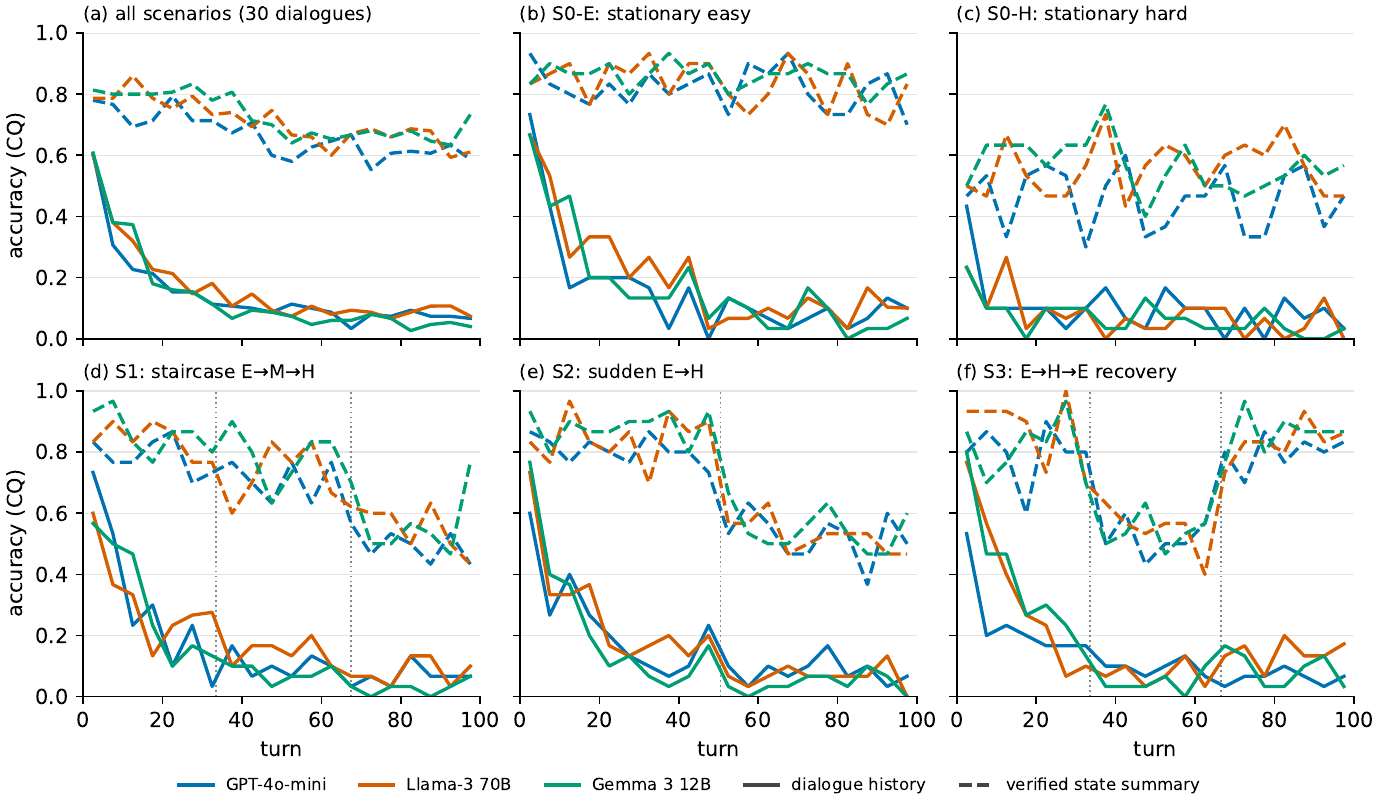}
\caption{Study 2 accuracy over 100 turns in 5-turn bins: (a) all 30 dialogues
per model; (b)--(f) the five scenarios, six dialogues per model each. Solid:
dialogue history; dashed: verified state summary; dotted vertical lines:
scheduled difficulty changes.}
\label{fig:s2}
\end{figure}

\paragraph{Accuracy collapses under accumulated history}
In all three models, dialogue-history accuracy fell from about 0.6 in the first
five turns to about 0.1 after turn 20 (Table~\ref{tab:s2collapse},
Figure~\ref{fig:s2}a). The curves of the three models were nearly identical
despite different sizes and mappings, and the collapse appeared in every
scenario. In stationary easy dialogues, where the questions never became
harder, accuracy over turns 21--100 was \cdrSzeroEDialRange{} under dialogue
history against \cdrSzeroECtrlRange{} under the verified state summary, both
restricted to S0-E (Figure~\ref{fig:s2}b); Table~\ref{tab:s2collapse}'s state-summary
column is pooled over all five scenarios, not S0-E alone, so the two numbers
are not read from the same column. At the
same ladder level, late turns were less accurate than early turns by
\cdrLevelMatched{} (prespecified secondary, $p$~\cdrLevelMatchedP{},
$n=\cdrLevelMatchedN$ dialogues; \cdrLevelMatchedSzeroE{} in S0-E). Because most
of the decline occurs in the first half, this contrast understates the total
loss. The collapse also hides the scheduled task drifts: it is present in
stationary dialogues and begins before any detector could form a baseline, so
pre/post contrasts around the scheduled changes mix the two (supplementary
Table~S7).

\begin{table}[!htbp]
\centering
\caption{Study 2 accuracy (CQ) by turn range under dialogue history, pooled and
in the stationary easy scenario (S0-E), and under the verified state summary
(turns 21--100). Thirty dialogues per model.}
\label{tab:s2collapse}
\small
\begin{tabular}{lccccc}
\toprule
 & \multicolumn{3}{c}{Dialogue history, turns} & S0-E, & State summary,\\
\cmidrule(lr){2-4}
Model & 1--5 & 6--20 & 21--100 & turns 21--100 & turns 21--100\\
\midrule
GPT-4o-mini & 0.61 & 0.25 & 0.09 & 0.10 & 0.65 \\
Llama-3 70B & 0.60 & 0.31 & 0.11 & 0.14 & 0.69 \\
Gemma 3 12B & 0.61 & 0.31 & 0.08 & 0.10 & 0.71 \\
 
\bottomrule
\end{tabular}
\end{table}

\paragraph{A verified state summary preserves accuracy (confirmatory)}
Under the verified state summary, accuracy stayed high and followed the
difficulty schedule, stepping down at the changes of S1 and S2 and recovering in
the final easy regime of S3, a recovery that dialogue history never showed
(Figure~\ref{fig:s2}). H-C1 was supported for both members of the family
(Table~\ref{tab:s2control}): over turns 1--100 the gap was
\cdrLlamaControlGap{} for Llama-3 70B and \cdrGemmaControlGap{} for Gemma 3 12B
(Holm $p$~\cdrLlamaControlHolmP{} and \cdrGemmaControlHolmP{}), and over turns
21--100 it was \cdrLlamaControlLateGap{} and \cdrGemmaControlLateGap{}.
GPT-4o-mini showed the same direction.

\begin{table}[!htbp]
\centering
\caption{Frozen test H-C1: per-dialogue accuracy of the verified state summary
minus dialogue history over turns 1--100. GPT-4o-mini was observed before the
protocol and is outside the Holm family.}
\label{tab:s2control}
\small
\begin{tabular}{lcccc}
\toprule
Model & Mean gap & 95\% bootstrap CI & Holm $p$ & Dialogues\\
\midrule
GPT-4o-mini & 0.521 & [0.476, 0.566] & -- & 30 \\
Llama-3 70B & 0.545 & [0.513, 0.576] & $<$0.001 & 30 \\
Gemma 3 12B & 0.588 & [0.557, 0.619] & $<$0.001 & 30 \\
 
\bottomrule
\end{tabular}
\end{table}

\section{Can the degradation be detected early?}
\label{sec:monitoring}

A deployed system does not observe accuracy, so we asked whether the framework
of Section~\ref{sec:framework} would have warned. ``Stationary'' here is the
task-difficulty schedule (S0-E, S0-H), not the metric: a metric can still
drift on a stationary task, and that drift is the false-alarm risk a
calibrated detector must control. We separate three evaluations. (i) The
\emph{initial collapse}: under dialogue history, accuracy falls within the
first 20 turns even on stationary tasks, before any detector has a reference
period. (ii) \emph{Scheduled-drift detection}: under the verified-state arms,
which remove that background collapse, can a detector flag the
experimenter-imposed changes of S1--S3 within a bounded delay without alarming
on S0-E/S0-H? (iii) \emph{Relative timing}: where both a candidate signal and
CQ alarm, does the candidate alarm first, and does any such lead replicate
across models? No test passed any of these three.

\subsection{Quality metrics do not fall with accuracy}

\begin{figure}[!htbp]
\centering
\includegraphics[width=\linewidth]{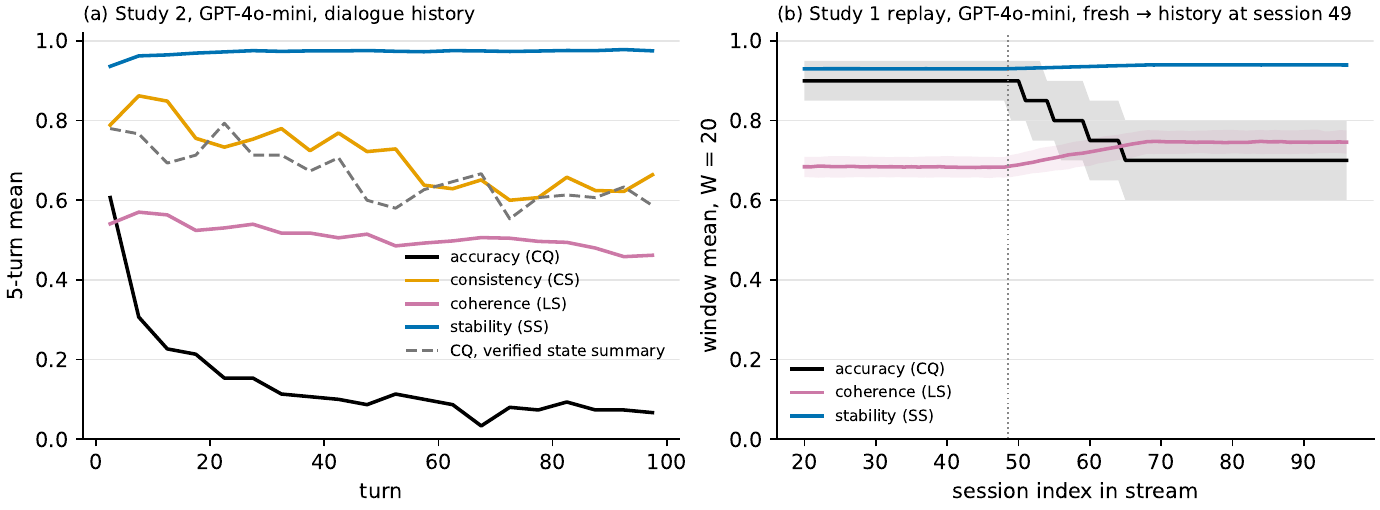}
\caption{Quality metrics do not track accuracy. (a) Study 2, GPT-4o-mini,
dialogue history, 5-turn means over 30 dialogues (descriptive); dashed:
accuracy under the verified state summary; CS is supported by
\silentCsSupport{} of 3,000 turns. (b) Post-hoc replay of Study 1: median and
10--90\% band over 1,000 orderings of window means ($W=20$) when the regime
switches from fresh prefix to dialogue history at session 49.}
\label{fig:silent}
\end{figure}

Coherence and stability moved in the wrong direction. In a post-hoc replay of
Study~1, in which completed sessions were streamed with a switch from fresh
prefix to dialogue history, both \emph{rose} while accuracy fell: at $W=20$ in
the sudden scenario, the median Cliff's $\delta$ \citep{cliff1993dominance}
between the last pre-change and the last window was \replaySsDeltaGpt{} for SS
in GPT-4o-mini, \replaySsDeltaLlama{} in Llama-4 Scout, and \replayLsDeltaGpt{}
for LS in GPT-4o-mini (Figure~\ref{fig:silent}b). Study~2 shows the same
pattern over 100 turns (Figure~\ref{fig:silent}a, descriptive). For GPT-4o-mini
under dialogue history, accuracy fell from \silentCQEarly{} (turns 1--5) to
\silentCQLate{} (turns 81--100), while stability rose from \silentSSEarly{} to
\silentSSLate{} and coherence moved only from \silentLSEarly{} to
\silentLSLate{}. Consistency fell from \silentCSEarly{} to \silentCSLate{}, but
slowly and long after most accuracy was lost; over turns 6--15 it even rose as
accuracy halved. As the model fails, its replies become more alike and remain
locally coherent, so a monitor that alarms on decreases cannot see the failure,
and a two-sided one would see a change whose sign says nothing about
correctness.

\subsection{Calibrated single-stream detectors}

In the Study~1 replay, with thresholds set on 2,000 permutation nulls at
family-wise $\alpha=0.05$ and checked on 2,000 disjoint nulls, the best
detection rate of any configuration was 0.067 (supplementary
Table~S3; there the stream items are sessions, and a detection is
any alarm after the switch). In the Study~2 dialogue arm the stationary
dialogues that should serve as nulls collapsed themselves, so for CQ no
threshold met the 5\% false-alarm target in most configurations, and the other
metrics detected at most 0.11 of drifts within 33 turns (supplementary
Table~S9). This is the central difficulty of monitoring under
accumulated history: the decline starts at turn~1, before any reference period,
in the very streams that define normal behaviour.

The verified-state arms remove this background collapse and leave only the
scheduled drift. Rules developed exploratorily on GPT-4o-mini and applied
unchanged to Llama-3 70B and Gemma 3 12B failed the frozen validation in both
models: CUSUM on the CQ window mean detected \valLlamaDetect{} and
\valGemmaDetect{} drift dialogues within 33 turns, and a baseline threshold
rule on LS signalled in \valLlamaLsSignal{} and \valGemmaLsSignal{} drift
dialogues but also in \valLlamaLsNull{} and \valGemmaLsNull{} stationary ones
(supplementary Section~S2.7). Our generic ADWIN baseline
never alarmed because the implementation omitted the variance term of the cut
bound of \citet{bifet2007learning}; re-running the affected analyses with a
corrected ADWIN and with CUSUM changed no conclusion (supplementary
Section~S4).

The frozen monitor forms a $W=20$ window and then ignores the first 20 window
values, so it cannot alarm before turn 40, after the first scheduled change at
turn 34. A post-hoc audit removed this floor with a CUSUM on raw values
standardised against turns 1--30 (supplementary Section~S5). On synthetic
streams it caught a shift injected at turn 31 by turn 33 in 45\% of streams for
a shift of 0.3 and in nearly all for 0.5, where the frozen monitor caught none.
Table~\ref{tab:audit} applies it to the control arms, with thresholds set on
GPT-4o-mini and transferred unchanged. The CQ detector did not transfer, and
LS, the label-free candidate, detected few drift dialogues and was usually
later than CQ. Its earlier alarms in GPT-4o-mini (3 of the 5 dialogues in which
both alarmed) were not repeated in the other models. ES, which needs the gold
answer, did no better (supplementary Table~S17). The result is therefore not
an absence of early alarms but low coverage, mostly late alarms and poor
transfer. It concerns the scheduled changes only; the early collapse under
dialogue history begins inside the audit's own reference period.

\begin{table}[!htbp]
\centering
\caption{Post-hoc fixed-baseline CUSUM on the control arms (18 drift and 12
stationary dialogues per model; thresholds set on GPT-4o-mini). LS first:
dialogues in which LS alarmed before CQ, among those in which both alarmed;
offset: median LS alarm turn minus CQ alarm turn.}
\label{tab:audit}
\small
\begin{tabular}{lcccccc}
\toprule
 & \multicolumn{2}{c}{Stationary alarms} & \multicolumn{2}{c}{Drift detected} & & \\
\cmidrule(lr){2-3}\cmidrule(lr){4-5}
Model & CQ & LS & CQ & LS & LS first & Offset\\
\midrule
GPT-4o-mini & 0/12 & 0/12 & 13/18 & 5/18 & 3/5 & $-1$\\
Llama-3 70B & 3/12 & 1/12 & 14/18 & 5/18 & 0/2 & $+27$\\
Gemma 3 12B & 2/12 & 0/12 & 15/18 & 8/18 & 1/7 & $+10$\\
\bottomrule
\end{tabular}
\end{table}

\subsection{Why single streams are underpowered}

A signal can fail to warn for two unrelated reasons: it may carry no
information about the coming failure, as CS, LS and ES did above, or it may
carry some information that a single 100-turn stream is simply too short to
detect at a controlled false-alarm rate. The bound and simulation below
quantify the second reason only, for CQ itself; they do not rehabilitate the
other metrics, whose failure was a direction and transfer problem, not a
sample-size one.

A simple bound shows how little one stream can reveal. Consider the accuracy
window mean (\ref{eq:windowmean}) pooled over $m$ dialogues at the same turn,
so that it averages $n=mW$ binary outcomes.

\begin{proposition}[Smallest detectable drop]
\label{prop:mindrop}
Let the $n=mW$ outcomes in the window be independent Bernoulli variables with
success probability $p_0$ before a change and $p_1=p_0-\delta$ after it, and
write $\sigma_i^2=p_i(1-p_i)$. A one-sided test that alarms when the window
mean falls below a threshold with false-alarm probability $\alpha$ under $p_0$
detects the drop with probability at least $\beta$, under the normal
approximation, only if
\begin{equation}
  \delta \;\ge\; \frac{z_{1-\alpha}\,\sigma_0 + z_{\beta}\,\sigma_1}{\sqrt{mW}},
  \label{eq:mindrop}
\end{equation}
where $z_q$ is the $q$-quantile of the standard normal distribution.
\end{proposition}

\begin{pf}
Under $p_i$ the window mean is approximately normal with mean $p_i$ and
standard deviation $\sigma_i/\sqrt{n}$. A false-alarm probability $\alpha$
requires the threshold $k=p_0-z_{1-\alpha}\sigma_0/\sqrt{n}$, and the detection
probability under $p_1$ is $\Phi\!\left((k-p_1)\sqrt{n}/\sigma_1\right)$. This
is at least $\beta$ exactly when $(k-p_1)\sqrt{n}\ge z_\beta\sigma_1$, which
gives (\ref{eq:mindrop}) after substituting $k$ and $p_0-p_1=\delta$.
\end{pf}

With $p_0=0.8$, $W=20$, $\alpha=0.05$ and $\beta=0.8$, the bound is about 0.24
for one dialogue, 0.10 for six and 0.05 for 18. A monitor that looks at many
turns needs a smaller per-look level; splitting $\alpha$ over the 67 looks from
turn 34 to 100 raises the bound to 0.38, 0.15 and 0.09, and positive dependence
between turns raises it further. The proposition covers only a threshold on
the window mean, so it explains the scale and the $1/\sqrt{m}$ gain from
pooling rather than the exact power of CUSUM or the other detectors.

\begin{figure}[!htbp]
\centering
\includegraphics[width=\linewidth]{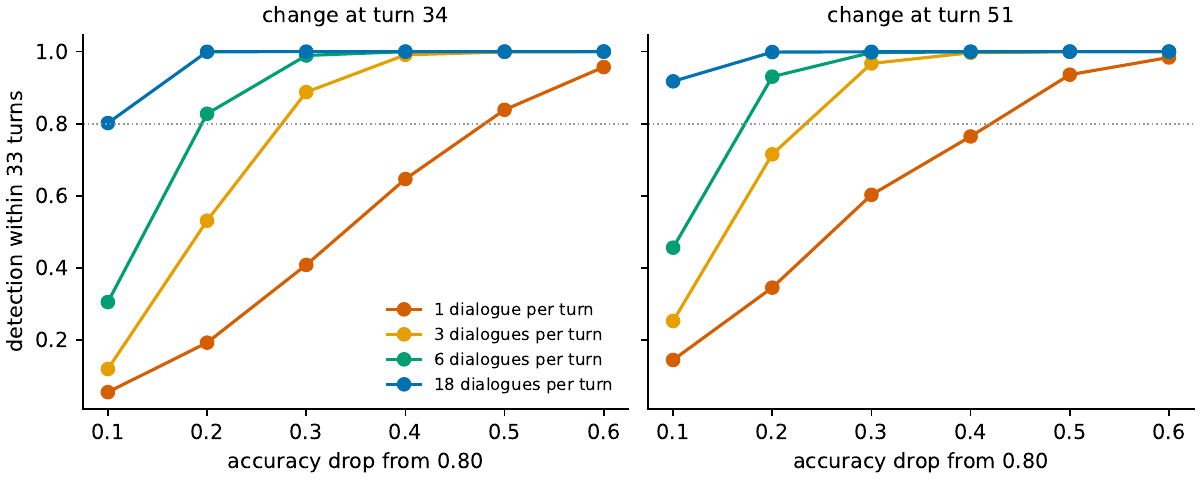}
\caption{Simulated detection within 33 turns versus the accuracy drop from 0.80
and the number of dialogues pooled per turn ($W=20$, thresholds from 12 null
streams, 5\% false-alarm target). Left: change at turn 34; right: at turn 51.
Dotted line: 80\% detection.}
\label{fig:power}
\end{figure}

For the detectors actually used, we simulated correctness streams with
pre-change accuracy 0.80 and a drop at turn 34 or 51, pooling $m$ dialogues per
turn, with thresholds from 12 null streams at a 5\% false-alarm target
(Figure~\ref{fig:power}). A single stream reached 80\% detection within 33
turns only for drops of about 0.5, whereas the task drifts in our data were
0.1--0.3; six dialogues per turn lowered the detectable drop to 0.2, and 18 to
0.1, with similar results for other windows and calibration sets
(supplementary Table~S13). On the control arms, a CQ CUSUM on
six-dialogue aggregates detected all three drift scenarios in both validation
models, with \valLlamaAggNull{} and \valGemmaAggNull{} alarms in the stationary
Llama and Gemma aggregates. Pooling sessions is therefore promising, but two
null aggregates per model cannot verify a 5\% false-alarm rate: with no alarm
in $n$ null streams, the exact one-sided 95\% upper bound is $1-0.05^{1/n}$,
which is 0.78 for $n=2$ and 0.22 for $n=12$.

\section{Discussion}
\label{sec:discussion}

\paragraph{What the two studies show}
Study~1 isolates the net cost of keeping the model's own replies, with the user
information held fixed. Study~2 shows a far larger loss in long state tracking,
but it does not separate history length, the number of state updates and
accumulated errors, because the verified summary changes all three at once.
Its result is therefore a joint failure of state tracking under a growing
transcript, not evidence that the assistant's replies alone cause the collapse;
that narrower claim rests on Study~1. Either way, a raw transcript as the only
memory of a stateful task has a large and avoidable cost.

\paragraph{Why the quality metrics stay silent}
Consistency, coherence and stability describe local properties of responses,
and a model that has adopted a wrong state can reason coherently and repeatably
from it. In our data these metrics fell far less than accuracy, and stability rose,
which
suggests that accumulated context makes responses more anchored rather than
more erratic, in line with reports of self-conditioning and sycophancy
\citep{sinha2025illusion,sharma2024sycophancy}. Two classes of signal remain
untested. Token entropy, used by ERGO \citep{khalid2025ergo}, needs token
probabilities, which our protocols did not collect. Semantic entropy
\citep{farquhar2024detecting} can be computed from sampled answers alone; our
consistency metric also summarises agreement among samples, but it is a
pairwise agreement rate rather than an entropy, and we did not compute
semantic entropy. Counterfactual signals, which compare the deployed context
with a history-free or state-grounded re-query, target anchoring directly;
CRL tracked the current readiness gap rather than anticipating it, and
comparing answers rather than readiness is a natural next step.

\paragraph{Implications for LLM-based information systems}
Where the task state can be represented explicitly, as in ledgers,
inventories, protocols and schedules, it should be kept in a verified store and
sent with each request; in our setting this kept accuracy 0.52--0.59 above full
history and shrank the prompt at turn 100 from tens of thousands of tokens to a
few hundred; summaries written by the model itself should be expected to do
less well than our solver-verified one. A dashboard of one conversation's
consistency, coherence or stability is likely to stay reassuring while accuracy
collapses, so pooling sessions or comparing against counterfactual re-queries
is the better route. Any detector should first be shown to fire on streams of
the intended length, with a reference period free of the degradation it is
meant to detect.

\paragraph{Limitations}
Both studies use arithmetic and state-tracking tasks with exact answers, which
make accuracy observable but are narrower than open-ended dialogue. All models
were accessed through one academic gateway, the two studies use different
model sets, and the models are not the strongest available. This is partly a
property of the design: calibration needs an informative accuracy range, and
a stronger open model, Gemma~4 31B, stayed above the hard band even at the
contingency levels of the ladder (supplementary
Section~S2.3), so testing frontier models requires a
harder ladder rather than only more calls. In Study~2, the repeat-based
metrics were complete only for GPT-4o-mini, and no user-only arm was run, so
the share of the collapse due to the assistant's replies is not estimated.
Monitoring analyses beyond the frozen tests are exploratory or post hoc, two
deviations for Llama-3 70B followed gate results and are disclosed, and one
investigator performed the task audit and calibration.

\section{Conclusion}
\label{sec:conclusion}

Keeping an LLM's own replies lowered success on short multi-turn tasks, and in
100-turn state tracking the full history collapsed accuracy within about twenty
turns at matched difficulty, while a verified state summary preserved it. The
loss was silent to the monitors we tested: no quality metric, paired readiness
measure or calibrated single-stream detector gave robust, model-general early
warning; coherence changed little and stability rose as accuracy fell.
Systems built on LLMs should keep task state outside the transcript and monitor
across sessions or against counterfactual re-queries. Open questions include
uncertainty-based and counterfactual signals, the separate roles of history
length, state load and error accumulation, and state files that models keep
and verify themselves.

\section*{Data and code availability}
All prompts, frozen protocols, configurations, analysis code and the scripts
that generate every number, table and figure in this article are publicly
available at \url{https://github.com/senolali/reasoning-monitor-3501}. 

\section*{Acknowledgements}
This work was supported by the Scientific and Technological Research Council of
T\"urkiye (T\"UB\.{I}TAK) under the 3501 programme, project number 126E663.

\section*{Declaration of competing interest}
The author declares that he has no known competing financial interests or
personal relationships that could have appeared to influence the work reported
in this paper.

\section*{CRediT authorship contribution statement}
Ali \c{S}enol: Conceptualization, Methodology, Software, Validation, Formal
analysis, Investigation, Data curation, Visualization, Writing--original draft,
Writing--review and editing, Project administration, Funding acquisition.

\section*{Declaration of generative AI and AI-assisted technologies in the manuscript preparation process}
During the preparation of this work the author used Claude (Anthropic) in an assistive capacity, for analysis code, figure scripts and language editing. The author designed the studies, carried out the research and reviewed and edited all resulting content, and takes full responsibility for the content of the published article. The language models evaluated in Studies 1 and 2 are objects of study, not writing aids.

\bibliographystyle{elsarticle-harv}
\bibliography{references}

\end{document}


\sloppy
\maketitle

\noindent This supplement gives design details, audit trails and secondary
results. Each result carries the evidence label defined in the article
(confirmatory, prespecified secondary, exploratory or post hoc); sensitivity
analyses are reported next to frozen results, never in their place. All tables
are generated by scripts in the public repository from hash-checked inputs.

\section{Study 1}

\subsection{Sample, prompts and parsing}
The public math subset of Lost in Conversation has 103 tasks and 585 shards.
Before any model call, eleven tasks were excluded because their normalised
question matched an input of our earlier experiments, and eight because their
shard sequence appeared to alter the problem; two tasks fell into both groups,
leaving 86. A hash-based draw that used no model output selected four
integration tasks, used only to test transport, resumption, parsing, scorers
and plotting, and \nTasks{} science tasks.

The system prompt read: \emph{``Solve incrementally using only information
supplied in this conversation. Prior assistant replies may be wrong and are
not authoritative facts.''} The instruction asked for concise numbered steps
ending with exactly one final line, \texttt{FINAL\_ANSWER: <number>} or
\texttt{STATUS: NEED\_MORE\_INFORMATION}; the two rewordings changed the wording
but not this contract. Both arms received a packed transcript in one user
message. Because the strict final-line parser rejected many replies that did
carry a marker during integration, a semantic-contract parser that still
requires the explicit marker was frozen before the science run; raw responses
and strict scores are retained. Six empty Mistral-7B primaries were kept as
invalid, never retried, and represented in later history by
\texttt{[EMPTY\_PRIMARY\_RESPONSE]}.

\subsection{LS and SS: splitting, scoring and missingness}
\label{sec:ls-ss-detail}
Before any metric is computed, a completion is split into reasoning segments by
a structure-aware rule shared across both studies: paragraphs are the primary
unit, a standalone display-math paragraph is attached to the explanation before
it, and sentence splitting is used only when the whole reply is one
unstructured paragraph; the final-answer line is removed first and never
scored as a step. LS then scores every consecutive pair of segments with the
NLI cross-encoder (main text, Section~3) and averages one
minus the contradiction probability over all pairs in a reply; a single-segment
reply has no pair and is reported as missing, not as 1. SS scores pairs among
the $K$ sampled \emph{replies} to the same context, not within one reply, so
two samples that restate the same wrong answer in the same template score as
highly similar; this is the mechanism behind the rising SS trajectories of
Section~6 of the article. Both metrics are denominated over only the pairs
with two parseable sides; a reply with no parseable text contributes no pair
and does not lower the mean, which is why support is always reported alongside
the value.

\begin{table}[h]\centering\small
\caption{Study 1 primary endpoint restricted to tasks with format-valid final
responses in both arms (sensitivity analysis).}
\label{tab:formatvalid}
\begin{tabular}{lccc}\toprule
Model & Tasks retained & Paired difference & McNemar $p$\\\midrule
\inputtable{generated/table_format_valid.tex}
\bottomrule\end{tabular}\end{table}

\subsection{Cumulative Resolution Lag}
Table~\ref{tab:crl} gives the frozen per-model results. In the post-hoc
persistence check, the last pre-final readiness gap alone reached AUROC
\crlLastGapAuroc{}, and CRL within its strata \crlCondAuroc{}.

\begin{table}[h]\centering\small
\caption{Frozen CRL discrimination of a positive final readiness gap
(confirmatory, auxiliary). Cases/controls: units with and without the target.}
\label{tab:crl}
\resizebox{\linewidth}{!}{%
\begin{tabular}{lcccccc}\toprule
Model & Units & Cases/controls & Mean CRL & AUROC [95\% CI] & $p$ & $p_{\text{Holm}}$\\\midrule
\inputtable{generated/table_crl.tex}
\bottomrule\end{tabular}}\end{table}

\subsection{Per-model metric trajectories}
\begin{figure}[htbp]\centering
\includegraphics[width=0.85\linewidth]{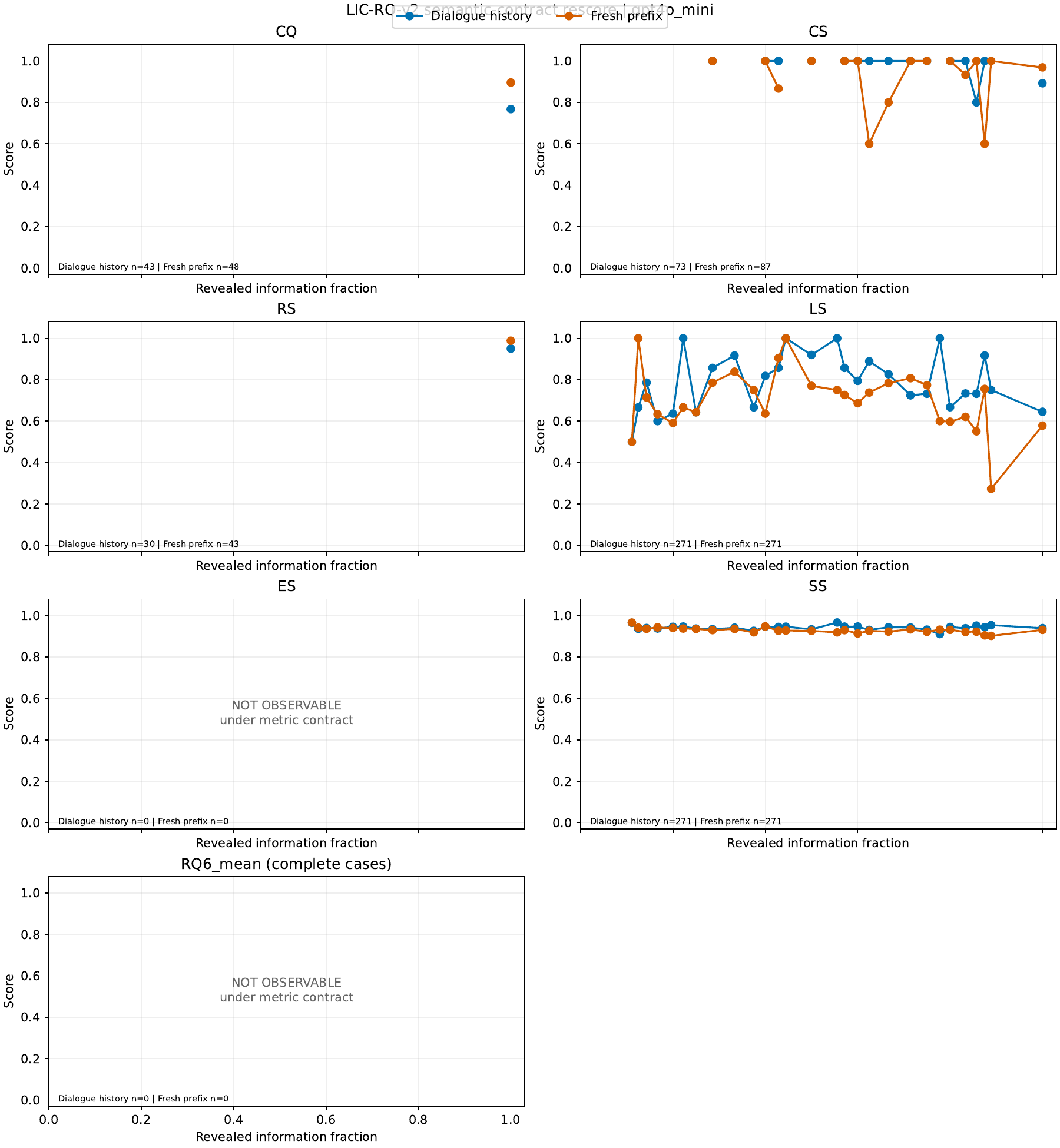}
\caption{Study 1, GPT-4o-mini: per-turn trajectories of the six dimensions and
their complete-case mean.}\label{fig:s1gpt}\end{figure}

\begin{figure}[htbp]\centering
\includegraphics[width=0.85\linewidth]{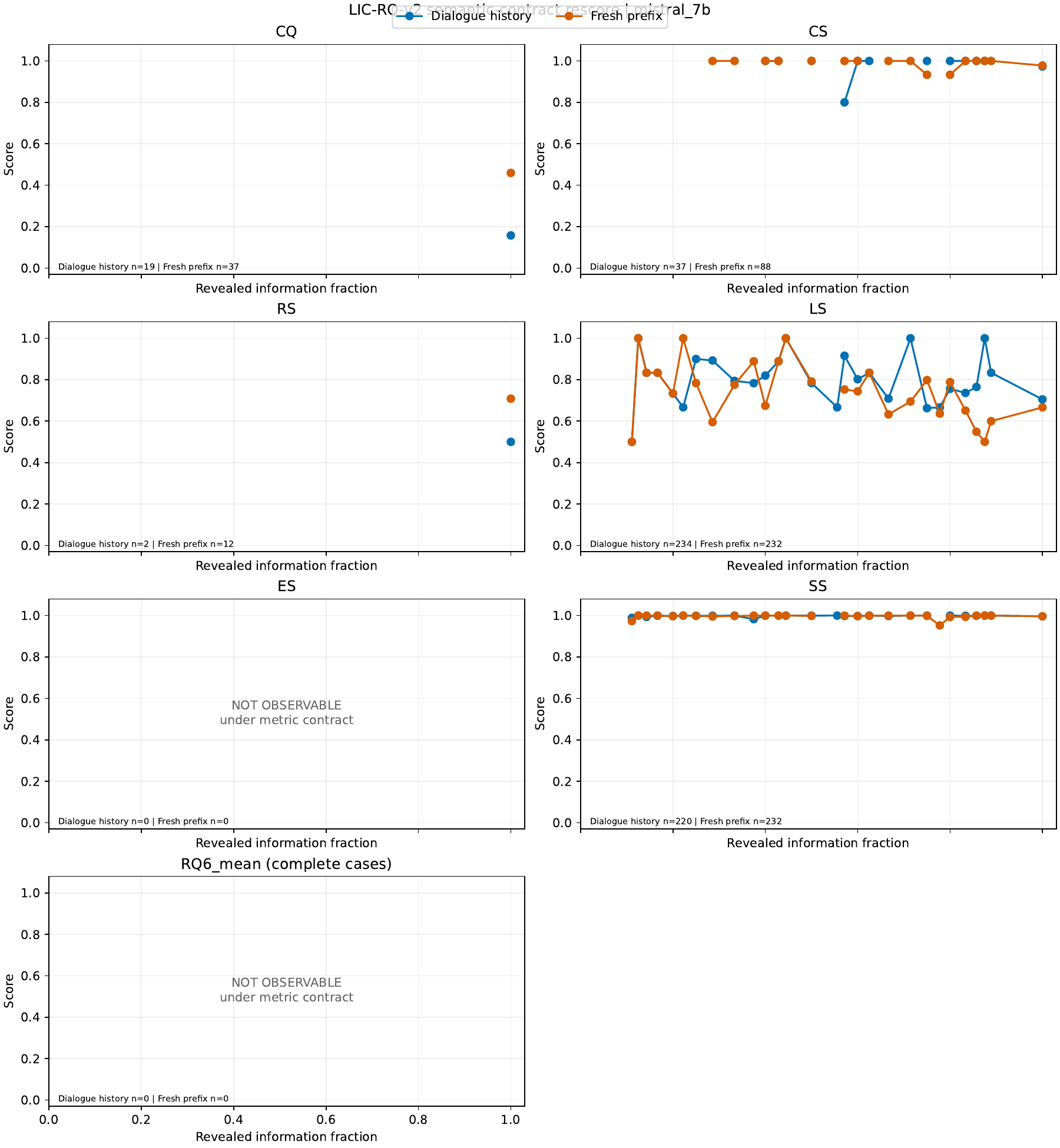}
\caption{Study 1, Mistral-7B.}\label{fig:s1mistral}\end{figure}

Figures~\ref{fig:s1gpt}--\ref{fig:s1mistral} show all six dimensions per turn
with their support. Gaps are structurally unobservable values, not zeros.

\begin{figure}[htbp]\centering
\includegraphics[width=0.85\linewidth]{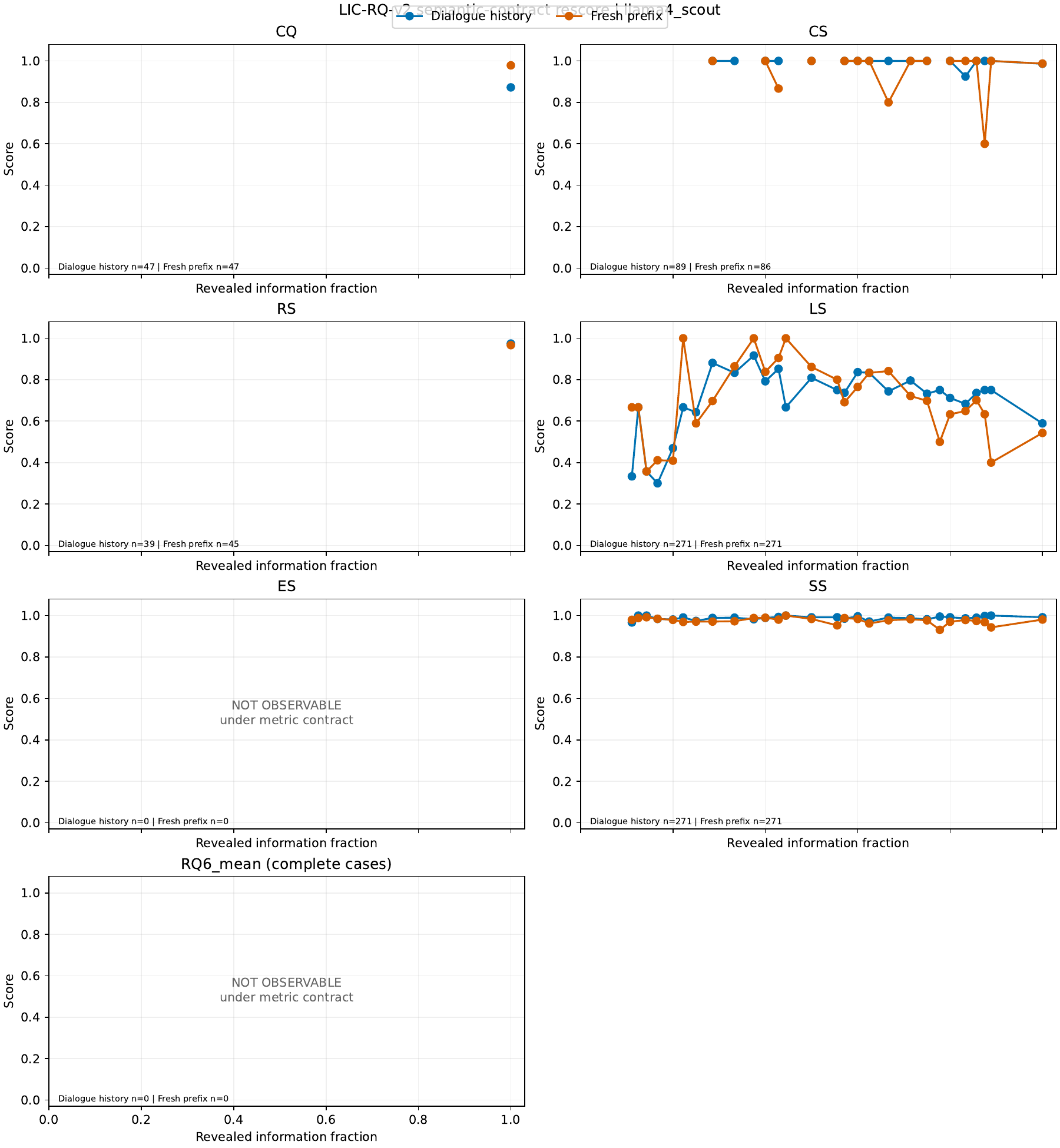}
\caption{Study 1, Llama-4 Scout.}\end{figure}

\subsection{Post-hoc stream replay}
Completed sessions were replayed as stream items with the regime switching
from fresh prefix to dialogue history, suddenly or gradually, over the same 48
tasks, with $W\in\{10,20,30\}$, five detectors and three assignment strategies.
Thresholds were set on 2,000 permutation nulls at family-wise $\alpha=0.05$ and
checked on 2,000 disjoint nulls. Table~\ref{tab:replay} gives the highest
detection rate per model and scenario. Over 180,000 replay rows, the historical
ADWIN never alarmed and the corrected ADWIN and CUSUM alarmed five times each,
with no early-warning candidate (Section~\ref{sec:adwin}).

\begin{table}[h]\centering\small
\caption{Study 1 replay at $W=20$ (post hoc): highest detection rate over the
validated cells.}
\label{tab:replay}
\begin{tabular}{llcl}\toprule
Model & Scenario & Max.\ detection & Metric / strategy\\\midrule
\inputtable{generated/table_replay.tex}
\bottomrule\end{tabular}\end{table}

\section{Study 2}

\subsection{Prompt and message construction}
\label{sec:s2-prompt}
Every call in both arms uses the same frozen system prompt -- \emph{``You keep
exact track of an evolving record for a user. Use only the information
provided in the message. Compute every quantity exactly.''} -- and the same
response contract -- \emph{``RESPONSE CONTRACT: Give at most eight short
numbered reasoning lines. Then end with exactly one final line in the form
FINAL\_ANSWER: <integer>.''} The dialogue-history user message packs the full
prior transcript as JSON (oldest first, with role) before the new turn and
contract; the verified-state message instead gives the solver's current
record for the affected entities, as text, before the same new turn and
contract (worked example: main text, Section~3).

\subsection{Generator, ladder and mappings}
\label{sec:s2-generator}
Each user turn realises one or more state events, may add distractor
sentences, and asks one question. Within a regime, 50\% of turns come from the
centre level, 20\% from each neighbouring level and 10\% from an adjacent
regime. Exact solvers keep the state and compute the
gold; the generator checks a solver round trip at every turn and that reworded
questions re-derive the same answer. Table~\ref{tab:ladder} gives the ladder, in
which exactly one axis changes per step; levels 13--14 are contingency levels
used only under the frozen ceiling rule. Table~\ref{tab:s2map} gives the frozen
mappings.

\begin{table}[h]\centering\small
\caption{Frozen difficulty ladder. Axes: query terms, events per turn,
distractor sentences and event complexity. $\dagger$: contingency levels.}
\label{tab:ladder}
\begin{tabular}{ccccc}\toprule
Level & Query terms & Events & Distractors & Complexity\\\midrule
\inputtable{generated/supp_ladder.tex}
\bottomrule\end{tabular}\end{table}

\begin{table}[h]\centering\small
\caption{Frozen difficulty mappings: ladder levels for the easy, medium and
hard regimes (E/M/H). A dash marks a domain excluded by the calibration rule.}
\label{tab:s2map}
\begin{tabular}{lcccc}\toprule
Model & Budget & Inventory & Laboratory & Milestones\\\midrule
\inputtable{generated/table_cdr_mappings.tex}
\bottomrule\end{tabular}\end{table}

\subsection{Difficulty calibration and model selection}
\label{sec:s2-calibration}
The frozen calibration rule fitted accuracy by isotonic regression and required
a one-sided Cochran--Armitage trend test with $p\le0.01$, adjacent-regime gaps of
at least 0.10 and an easy--hard gap of at least 0.25; a domain failing any
condition was excluded. A model was excluded if contract validity fell below
0.90 or any response was truncated. Over several frozen rounds, a coarse first
ladder was replaced by the one-axis-per-step ladder and a scheduling domain by
the milestone domain. Candidate models were replaced before any stream result
existed: Mistral-7B sat at the accuracy floor and truncated responses, the
gateway truncated a frontier model's outputs in one domain, and Gemma~4 31B
stayed at the ceiling even at the contingency levels and was replaced by Gemma
3 12B. Consistency and stability need diverse repeats. Llama-4 Scout produced
near-identical samples, and among the Llama variants only Llama-3 70B met the
frozen 0.80 diversity gate (Table~\ref{tab:kdiv}); a probe confirmed that it
accepts the full 100-turn history (37,139 prompt tokens) and recalls its oldest
message.

\begin{table}[h]\centering\small
\caption{Raw-text repeat diversity of Llama-family models on identical prompts.
Gate: diversity $\ge0.80$, contract validity $\ge0.90$, no truncation.}
\label{tab:kdiv}
\begin{tabular}{llccc}\toprule
Model & Prompts & Diversity & Validity & Truncated\\\midrule
\inputtable{generated/supp_kdiv.tex}
\bottomrule\end{tabular}\end{table}

\subsection{Deviations, acquisition and staged analysis}
Two deviations for Llama-3 70B were recorded after the relevant gate results.
One of its 1,440 calibration responses reached the 2,048-token ceiling, in a
domain excluded for this model on accuracy grounds, and the any-truncation rule
was overridden for this model only; Mistral-7B had been excluded under the same
rule with five truncations. Its technical pilot also had two truncated
non-primary responses, both repetition loops, and the pilot's zero-truncation
gate was overridden. In its place a truncation policy was frozen for all models
before the main run: a truncated reply is kept verbatim and scored as missing,
a truncated primary enters later history as a placeholder, and a sensitivity
analysis rescores truncated cells as incorrect.

The budget paused the main run twice. At the second pause, 25,039 of 54,000
calls had been made; all 90 dialogue-history primary chains were complete
except one Llama-3 70B turn-100 primary. The analysis protocol was frozen
before any main-run metric was computed. Stage A analysed every fully acquired
model--metric pair; Stage B, covering the remaining repeat-based samples and
controls for Llama-3 70B and Gemma 3 12B, was committed before any Stage A
result was seen. Its control arms (3,000 calls per model) were later completed
under a separately frozen protocol (test H-C1 in the article); its repeat-based
samples were not. At Stage A, Llama-3 70B had two truncated primaries, and
scoring them as incorrect changed no result.

\begin{figure}[htbp]\centering
\includegraphics[width=0.8\linewidth]{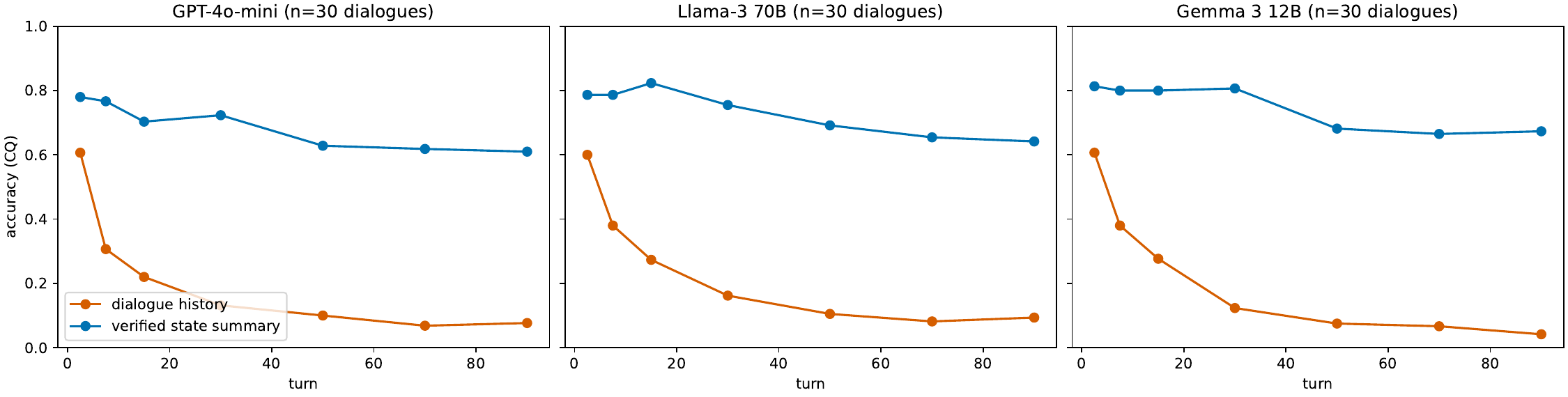}
\caption{Dialogue-history and verified-state-summary accuracy over turns for
the three Study 2 models.}
\label{fig:s2controlcompletion}
\end{figure}

\subsection{Frozen change-point contrasts}
The confirmatory Stage A contrasts compare the 20 turns after each change point
with the 20 turns before it (Table~\ref{tab:h1a}). Because the dialogue arm
collapses from turn 1, even in stationary dialogues, they mix task drift with
the background collapse and are not interpreted as task-drift effects. The
dialogue-minus-control contrast H1b, available for GPT-4o-mini only, was
uninformative under this floor (mean differences 0.03--0.23, Holm $p=1.0$).

\begin{longtable}{llcccc}
\caption{Stage A H1a (confirmatory): post-minus-pre window means in the
dialogue arm, pooled over models with model-stratified sign flips. CS, RS and
SS are GPT-4o-mini only.}
\label{tab:h1a}\\
\toprule Transition & Metric & Mean $\Delta$ & Dialogues & $p$ & $p_{\text{Holm}}$\\\midrule
\endfirsthead
\toprule Transition & Metric & Mean $\Delta$ & Dialogues & $p$ & $p_{\text{Holm}}$\\\midrule
\endhead
\inputtable{generated/supp_h1a.tex}
\bottomrule
\end{longtable}

\subsection{Detector strategies and dialogue-arm monitoring}
Table~\ref{tab:strategies} gives the three detector assignments and
Table~\ref{tab:stagea} the dialogue-arm detection rates; n/c means that no grid
value met the null false-alarm target, because the stationary dialogues also
decline.

\begin{table}[h]\centering\small
\caption{Detector assignment strategies. All monitor a decrease of the window
mean; CQ is additionally monitored on its raw binary stream by DDM.}
\label{tab:strategies}
\begin{tabular}{lcccccc}\toprule
Strategy & CQ & CS & RS & LS & ES & SS\\\midrule
Distribution-based & CUSUM & CUSUM & CUSUM & CUSUM & CUSUM & CUSUM\\
Proposal mapping & CUSUM & ADWIN & DDM & PH & EWMA & ADWIN\\
ADWIN everywhere & ADWIN & ADWIN & ADWIN & ADWIN & ADWIN & ADWIN\\
\bottomrule\end{tabular}\end{table}

\begin{table}[h]\centering\small
\caption{Stage A dialogue-arm detection rate within 33 turns of the first
change, with thresholds calibrated on the 36 pooled stationary dialogues. CQ,
LS and ES pool all three models; CS, RS and SS are GPT-4o-mini only, as in
Table~\ref{tab:h1a}.}
\label{tab:stagea}
\begin{tabular}{llcccccc}\toprule
$W$ & Strategy & CQ & CS & RS & LS & ES & SS\\\midrule
\inputtable{generated/supp_stagea_monitoring.tex}
\bottomrule\end{tabular}\end{table}

\subsection{Control-arm monitoring}
\label{sec:control-monitoring}
\begin{figure}[htbp]\centering
\includegraphics[width=\linewidth]{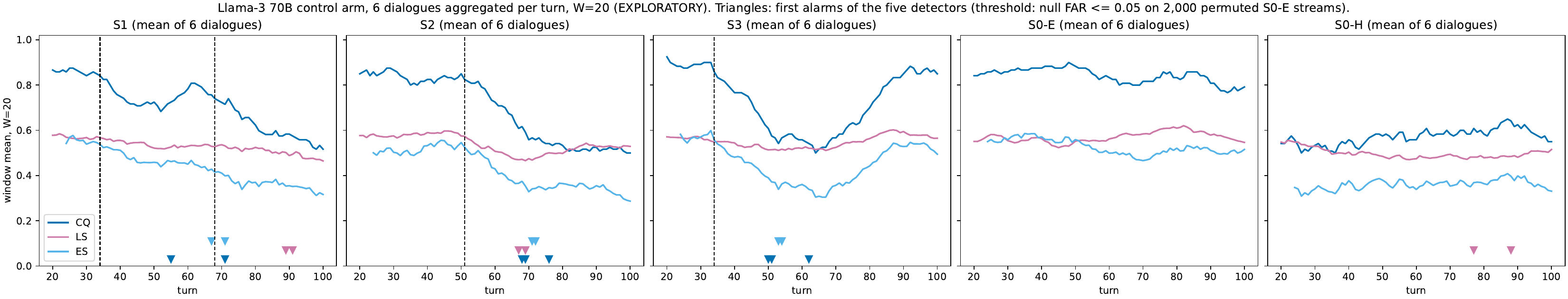}\\[4pt]
\includegraphics[width=\linewidth]{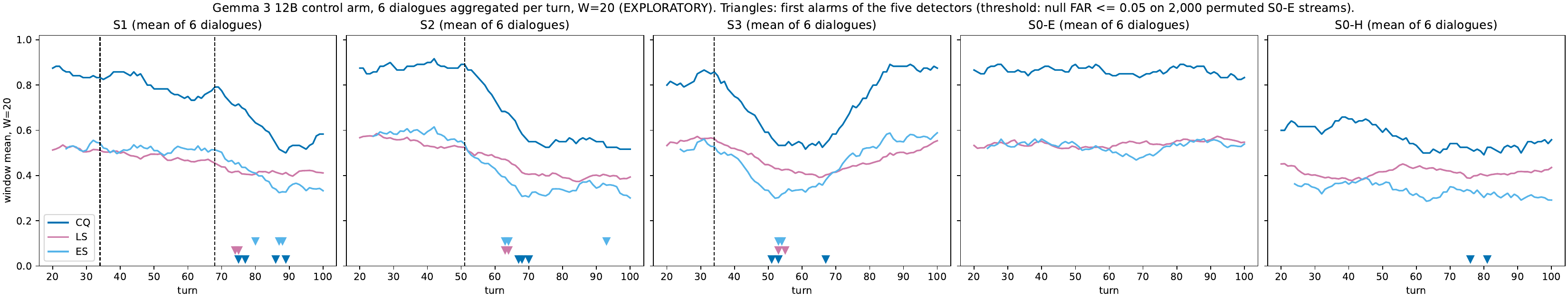}
\caption{Six-dialogue aggregate monitoring at $W=20$ in the frozen validation:
Llama-3 70B (top) and Gemma 3 12B (bottom).}\label{fig:valagg}\end{figure}

The rules were developed exploratorily on the GPT-4o-mini control arm
(Tables~\ref{tab:ctrlscreen} and~\ref{tab:early}) and then applied unchanged to
Llama-3 70B and Gemma 3 12B as a frozen validation (Table~\ref{tab:monitorvalidation},
Figure~\ref{fig:valagg}). On GPT-4o-mini, CQ CUSUM detected \ctrlScreenDetect{}
drift dialogues within 33 turns at $W=20$, and the baseline threshold rule
flagged accuracy in \earlyCqSignal{} drift but also \earlyCqNull{} stationary
dialogues; LS signalled in \earlyLsSignal{} drift dialogues, \earlyLsBefore{} of
them before accuracy, but also in \earlyLsNull{} stationary ones. Neither
validation model passed the single-stream screen, and the apparent LS leads did
not separate from alarms in stationary dialogues.

\begin{table}[h]\centering\small
\caption{GPT-4o-mini control arm (exploratory): per-dialogue detection rate
within 33 turns, with the leave-one-out false-alarm rate in parentheses, on 18
drift and 12 stationary dialogues.}
\label{tab:ctrlscreen}
\begin{tabular}{llccc}\toprule
$W$ & Strategy & CQ & LS & ES\\\midrule
\inputtable{generated/supp_control_screen.tex}
\bottomrule\end{tabular}\end{table}

\begin{table}[h]\centering\small
\caption{Per-dialogue threshold rule on GPT-4o-mini (exploratory). Baseline
turns 1--30; a signal is a window mean below $\mu_0-2\sigma_0/\sqrt{W}$ on three
consecutive turns. LS and ES timing relative to CQ: before/same/after.}
\label{tab:early}
\begin{tabular}{ccccccc}\toprule
$W$ & CQ signalled & CQ null & LS signalled & LS b/s/a & LS null & ES b/s/a\\\midrule
\inputtable{generated/supp_early_signal.tex}
\bottomrule\end{tabular}\end{table}

\begin{table}[h]\centering\small
\caption{Frozen monitoring validation at $W=20$. Null counts are alarms among
12 stationary dialogues for the LS rule and among two stationary scenario
aggregates for the six-dialogue CQ CUSUM.}
\label{tab:monitorvalidation}
\begin{tabular}{lcccccc}\toprule
Model & CUSUM detect & Delay & LS signal & LS null & Agg.\ detect & Agg.\ null\\\midrule
Llama-3 70B & \valLlamaDetect{} & \valLlamaDelay{} & \valLlamaLsSignal{} &
\valLlamaLsNull{} & \valLlamaAggDetect{} & \valLlamaAggNull{}\\
Gemma 3 12B & \valGemmaDetect{} & \valGemmaDelay{} & \valGemmaLsSignal{} &
\valGemmaLsNull{} & \valGemmaAggDetect{} & \valGemmaAggNull{}\\
\bottomrule\end{tabular}\end{table}

\subsection{Per-model trajectories (Stage A)}
\begin{figure}[htbp]\centering
\includegraphics[width=0.85\linewidth]{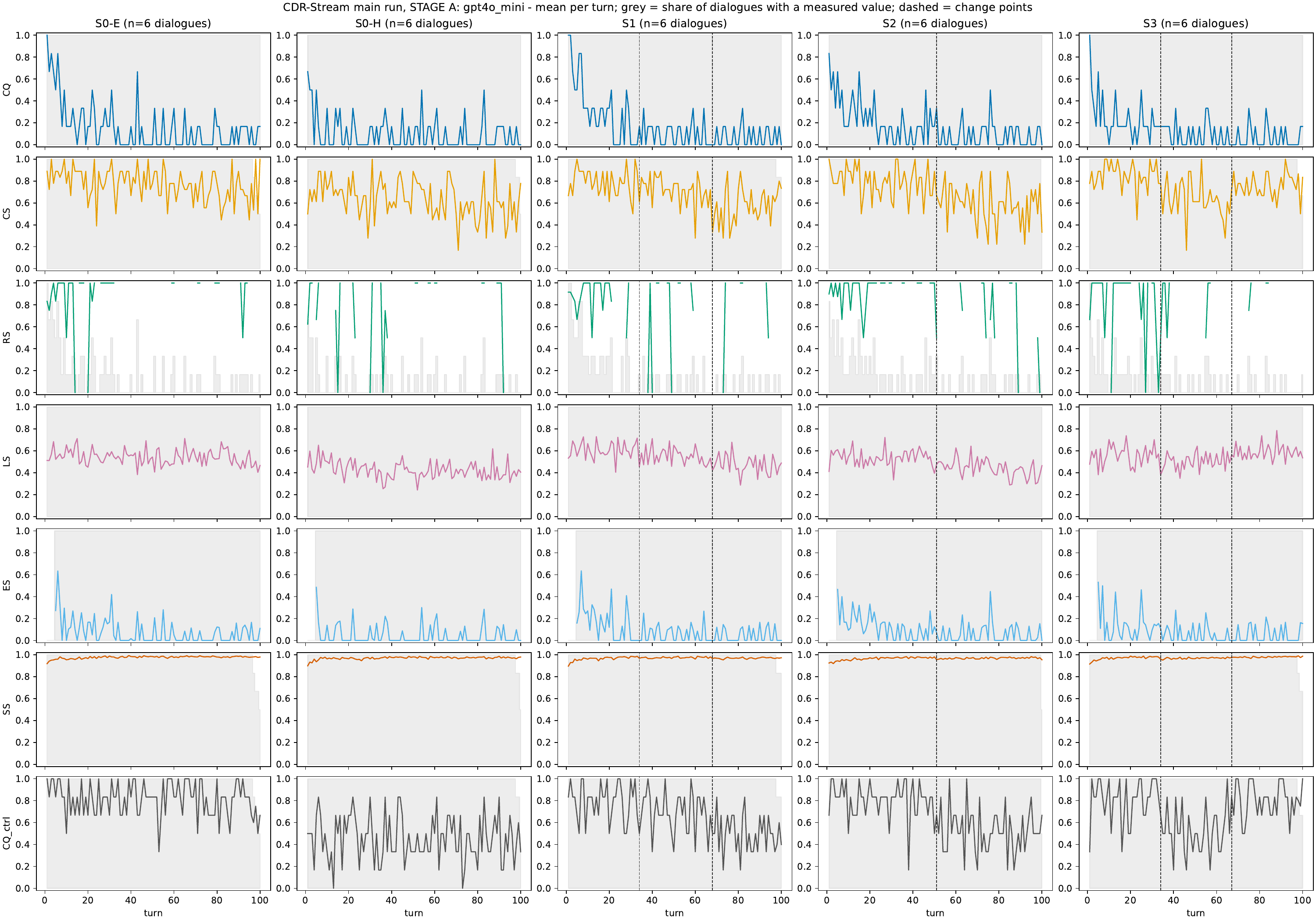}
\caption{Study 2, GPT-4o-mini: six dimensions and the control CQ by
scenario.}\end{figure}
\begin{figure}[htbp]\centering
\includegraphics[width=0.85\linewidth]{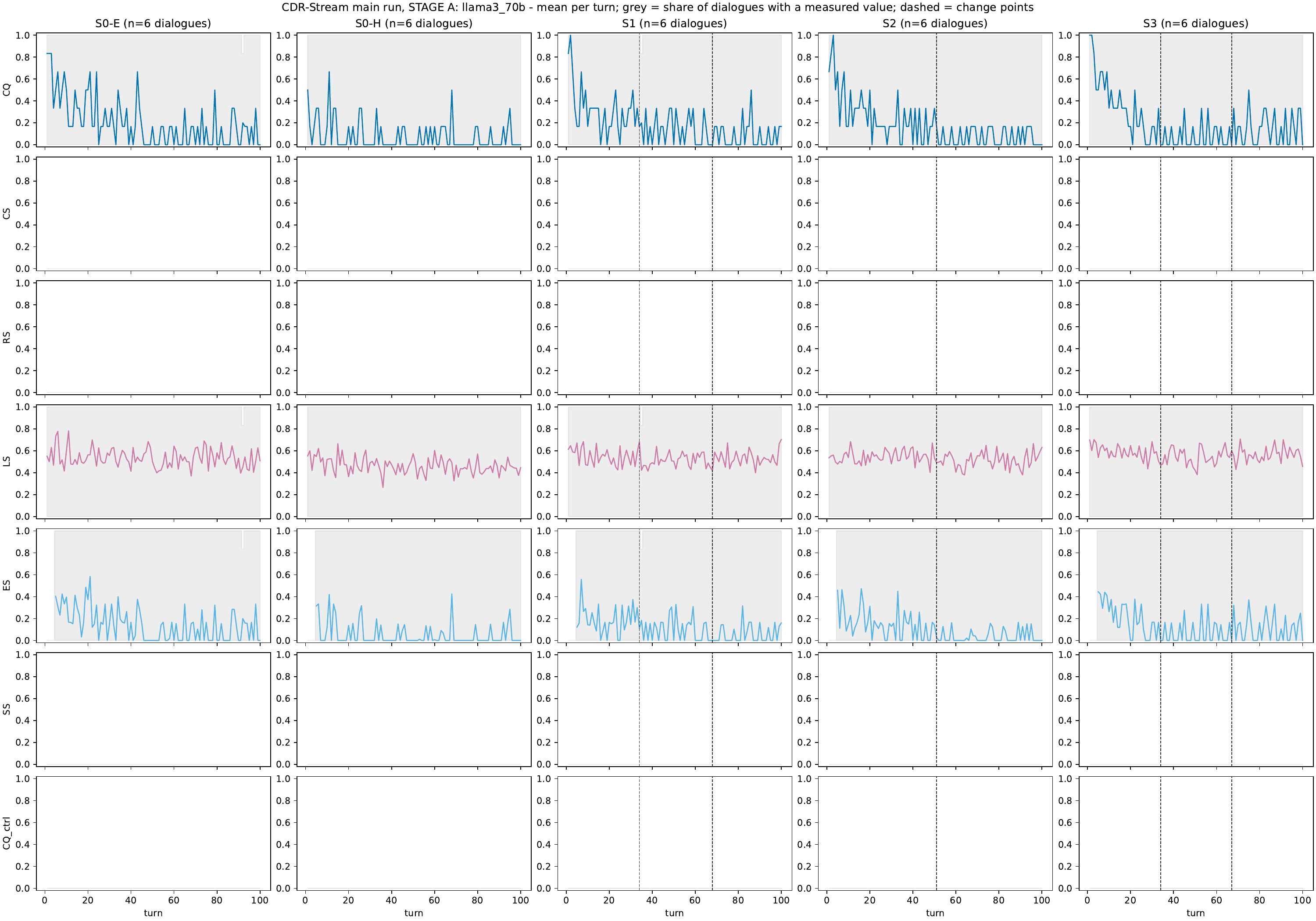}
\caption{Study 2 dialogue-arm metrics, Llama-3 70B. Repeat-based sampling was
not completed, so the CS, RS and SS panels are empty by design, not zero.}
\end{figure}
\begin{figure}[htbp]\centering
\includegraphics[width=0.85\linewidth]{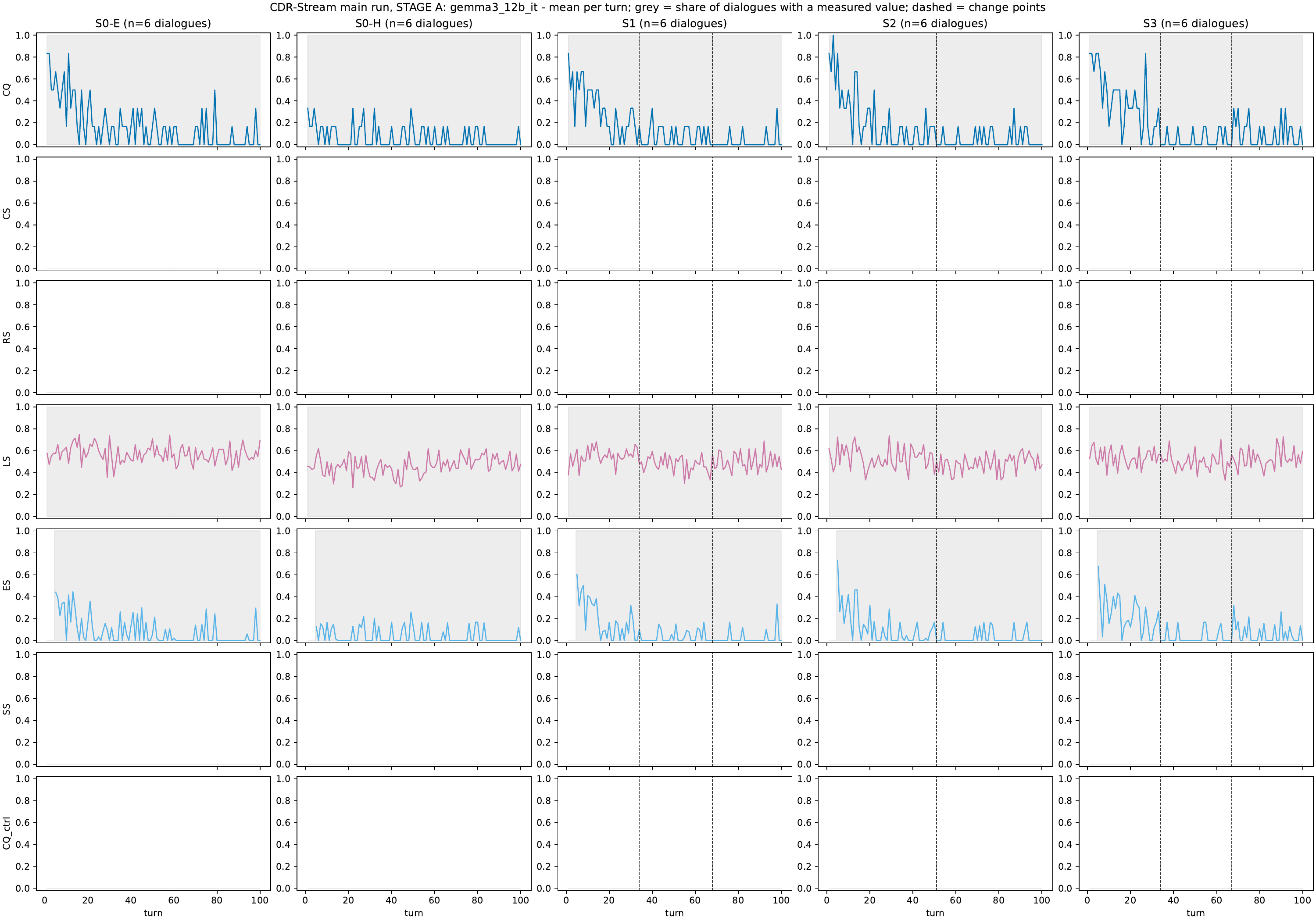}
\caption{Study 2 dialogue-arm metrics, Gemma 3 12B. Repeat-based sampling was
not completed, so the CS, RS and SS panels are empty by design, not zero.}
\end{figure}

\section{Detector power simulation}
Simulated streams have pre-change accuracy 0.80 and a drop $\delta$ at turn 34
or 51, with $m$ independent dialogues pooled per turn. The detectors and grids
are those used in the analyses. Thresholds were calibrated on 2,000 nulls
(oracle) or on 200 random subsets of 36 or 12 nulls, with 300 drift streams per
cell. Table~\ref{tab:powerfull} gives the smallest drop detected within 33
turns in at least 80\% of runs.

\begin{longtable}{lcccc}
\caption{Smallest detectable drop with the best detector, by calibration set,
window $W$ and dialogues per turn $m$.}\label{tab:powerfull}\\
\toprule Calibration & $W$ & $m$ & Change at 34 & Change at 51\\\midrule\endfirsthead
\toprule Calibration & $W$ & $m$ & Change at 34 & Change at 51\\\midrule\endhead
\inputtable{generated/supp_power_full.tex}
\bottomrule
\end{longtable}

\section{ADWIN implementation defect}
\label{sec:adwin}
The historical ADWIN used the cut bound $\epsilon=\sqrt{(2/m)\ln(2n/\delta)}$
without the variance term of \citet{bifet2007learning}. For $n\le100$ values in
$[0,1]$ this bound exceeds every attainable difference in means, so no ADWIN
cell in our tables could alarm. Every affected analysis was re-run with a
corrected ADWIN that uses the variance bound and with CUSUM in its place; no
recorded decision changed. The corrected ADWIN follows the published bound but
is too conservative for 100-turn streams (Table~\ref{tab:adwv}).

\begin{table}[h]\centering\small
\caption{Corrected ADWIN: detection by turn 100 ($W=20$, change at 51, oracle
calibration) by drop size and number of dialogues pooled per turn.}
\label{tab:adwv}
\begin{tabular}{ccccc}\toprule
$\delta$ & $m=1$ & $m=3$ & $m=6$ & $m=18$\\\midrule
\inputtable{generated/supp_adwin_v_power.tex}
\bottomrule\end{tabular}\end{table}

\section{Early-warning audit without the turn-40 floor (post hoc)}
\label{sec:forensic}
The frozen Study~2 monitor forms a $W=20$ window mean and lets each detector
ignore its first 20 window values, so it cannot alarm before turn 40, after the
first scheduled change at turn 34. This call-free audit, designed after the
frozen results were known, asks whether the negative result depends on that
floor. It changes no frozen result. The detector is a CUSUM on raw per-turn
values, standardised against a fixed baseline of turns 1--30 that later values
never update, with allowance 0.5 and no warm-up. Non-CQ metrics are monitored in
both directions with a shared threshold, because coherence or stability may
rise during confidently wrong reasoning. Thresholds are the smallest grid value
in $\{2,3,4,5,6,8,10,12,15,20\}$ with a stationary false-alarm rate of at most
0.05.

\paragraph{Synthetic precursor}
Streams of 100 values from $N(0.75,0.10^2)$, clipped to $[0,1]$, received a
downward shift $\delta$ from turn 31; thresholds were set on 2,000 null streams
and each $\delta$ was tested on 1,000 streams (Table~\ref{tab:forensic-synth}).
The frozen path never alarmed by turn 33, whereas the fixed-baseline CUSUM did
so in 45\% of streams for $\delta=0.3$ and almost always for $\delta=0.5$; both
detected shifts of 0.2 or more by turn 67.

\begin{table}[h]\centering\small
\caption{Synthetic shift injected at turn 31: share of streams with an alarm by
turn 33 and by turn 67. Null false-alarm rates on the calibration streams were
0.041 (fixed baseline) and 0.016 (frozen path).}
\label{tab:forensic-synth}
\begin{tabular}{lcccc}\toprule
& \multicolumn{2}{c}{Fixed-baseline CUSUM} & \multicolumn{2}{c}{Frozen path}\\
Shift $\delta$ & by turn 33 & by turn 67 & by turn 33 & by turn 67\\\midrule
0.0 & 0.000 & 0.021 & 0.000 & 0.004\\
0.1 & 0.002 & 0.931 & 0.000 & 0.759\\
0.2 & 0.053 & 1.000 & 0.000 & 1.000\\
0.3 & 0.447 & 1.000 & 0.000 & 1.000\\
0.5 & 0.995 & 1.000 & 0.000 & 1.000\\
\bottomrule\end{tabular}\end{table}

\paragraph{Control arms}
Thresholds were chosen on the GPT-4o-mini stationary dialogues and transferred
unchanged to Llama-3 70B and Gemma 3 12B, so the GPT-4o-mini false-alarm rates
are in-sample (Table~\ref{tab:forensic-emp}). The CQ detector did not transfer.
LS and ES stayed within the false-alarm target in most cases but detected under
half of the drift dialogues and rarely alarmed before CQ. In the GPT-4o-mini
dialogue-history arm, no threshold met the target for CS, RS or ES, and the
calibrated LS and SS rules detected 0.11 and 0 of the drift dialogues.

\paragraph{Prospective discrimination}
As a descriptive check, each metric was scored at turn $t$ by its standardised
deviation from the dialogue's own baseline of turns 1--30: the downward
deviation for CQ and the absolute deviation for the other metrics. Eligible
turns were turns 31 onward in the drift dialogues at which CQ was correct, and
the label was a CQ failure at turn $t+h$. We report the AUROC for the best
$h\in\{1,\dots,20\}$. On the control arms the best values were 0.52--0.57 on
616--780 eligible turns. In the GPT-4o-mini dialogue arm they were higher
(Table~\ref{tab:auc}), but almost every eligible turn there is followed by a
failure, so each value rests on 4--14 turns that are not, after choosing the
best of 20 horizons. The CQ row is a reference, not a warning signal: CQ equals
1 on every eligible turn, so its score is constant within a dialogue and its
AUROC only reflects differences in baseline accuracy between dialogues. These
values do not show that the metrics carry no predictive information, but they
cannot support a warning claim either. No confidence intervals are given,
because turns within a dialogue are dependent and the horizon was selected
post hoc.

\begin{table}[h]\centering\small
\caption{Best-horizon AUROC in the GPT-4o-mini dialogue-history arm
(descriptive, post hoc). Failure / no failure: eligible turns with and without
a CQ failure at turn $t+h$. For CQ the score is constant within a dialogue
(see text).}
\label{tab:auc}
\begin{tabular}{lcccc}\toprule
Metric & Best $h$ & AUROC & Failure at $t+h$ & No failure\\\midrule
CQ (reference) & 14 & 0.660 & 88 & 5\\
CS & 12 & 0.675 & 82 & 14\\
RS & 17 & 0.628 & 82 & 7\\
LS & 4 & 0.701 & 99 & 4\\
ES & 6 & 0.669 & 96 & 7\\
SS & 19 & 0.729 & 75 & 9\\
\bottomrule\end{tabular}\end{table}

\begin{table}[h]\centering\small
\caption{Fixed-baseline CUSUM on the control arms (18 drift and 12 stationary
dialogues per model). Earlier/both: dialogues in which the metric alarmed
before CQ, among those in which both alarmed. Offset: median metric alarm turn
minus CQ alarm turn (negative is earlier).}
\label{tab:forensic-emp}
\begin{tabular}{llcccc}\toprule
Model & Metric & Stationary FA & Detection & Earlier/both & Offset\\\midrule
GPT-4o-mini & CQ & 0.000 & 0.722 & -- & --\\
GPT-4o-mini & LS & 0.000 & 0.278 & 3/5 & $-1.0$\\
GPT-4o-mini & ES & 0.000 & 0.444 & 1/8 & 4.5\\
Llama-3 70B & CQ & 0.250 & 0.778 & -- & --\\
Llama-3 70B & LS & 0.083 & 0.278 & 0/2 & 27.0\\
Llama-3 70B & ES & 0.250 & 0.278 & 0/5 & 8.0\\
Gemma 3 12B & CQ & 0.167 & 0.833 & -- & --\\
Gemma 3 12B & LS & 0.000 & 0.444 & 1/7 & 10.0\\
Gemma 3 12B & ES & 0.000 & 0.389 & 1/7 & 3.0\\
\bottomrule\end{tabular}\end{table}

For the scheduled changes, the negative result therefore does not depend on
the turn-40 floor. The audit does not address the early collapse under
dialogue history, which begins inside its own reference period.

\section{Models, scorers and timeline}
\label{sec:timeline}

All models were called through a standard OpenAI-compatible chat-completions API, with server-side history, retrieval and prompt rewriting disabled. The API exposes aliases rather than dated vendor snapshots, so the aliases and acquisition dates in Table~\ref{tab:models} are the most precise version information available. Logical coherence used \texttt{cross-encoder/nli-deberta-v3-small} and stability used BERTScore with distilbert-base-uncased, both run locally. Table~\ref{tab:timeline} gives the order in which protocols were frozen and data collected.

\begin{table}[h]\centering\small
\caption{Models, gateway aliases and acquisition dates. All calls used
temperature 0.7 and a 2,048-token output ceiling.}
\label{tab:models}
\begin{tabular}{llll}\toprule
Study & Model & Gateway alias & Acquisition\\\midrule
1 & GPT-4o-mini & \texttt{openai/gpt4o\_mini} & 2026-10-02\\
1 & Llama-4 Scout & \texttt{llama4\_scout-17b} & 2026-10-02\\
1 & Mistral-7B & \texttt{mistral-7b} & 2026-10-02\\
2 & GPT-4o-mini & \texttt{openai/gpt4o\_mini} & 2026-10-03 to 2026-10-04\\
2 & Llama-3 70B & \texttt{llama3-70b} & 2026-10-03 to 2026-11-05\\
2 & Gemma 3 12B & \texttt{gemma3\_12b\_it} & 2026-10-03 to 2026-11-05\\
\bottomrule\end{tabular}\end{table}

\begin{table}[h]\centering\small
\caption{Dated timeline of protocols, acquisition and analyses.}
\label{tab:timeline}
\begin{tabular}{@{}p{0.14\linewidth}p{0.80\linewidth}@{}}\toprule
Date & Step\\\midrule
2026-10-01 & Study~1 protocol frozen: sample, prompts, models, scoring and
confirmatory analysis. Technical integration on four disjoint tasks (1,092
calls); no effect analysed.\\
2026-10-02 & CRL frozen. Study~1 acquisition (11,382 calls) completed, then the
confirmatory analysis run. Replay protocol frozen before any replay was
computed.\\
2026-10-03 & Study~2 calibration completed and mappings frozen. Main-run
sample, prompt, runner and truncation policy frozen; main acquisition
started.\\
2026-10-04 & Acquisition paused at 25,039 of 54,000 calls. Study~2 analysis
protocol frozen before any main-run metric was computed; Stage~A run.
Control-completion and control-monitoring protocols frozen. ADWIN audit.\\
2026-11-05 & Control arms for Llama-3 70B and Gemma 3 12B completed (3,000
calls each) and analysed under the frozen H-C1 protocol; frozen monitoring
validation run.\\
\bottomrule\end{tabular}\end{table}

\section{Reproducibility}
Every number, table and figure is produced by scripts from hash-checked inputs
(Table~\ref{tab:hash}). Raw responses are never overwritten.

\begin{table}[h]\centering\small
\caption{Key SHA-256 prefixes.}\label{tab:hash}
\begin{tabular}{ll}\toprule Artifact & SHA-256\\\midrule
\inputtable{generated/supp_hashes.tex}
\bottomrule\end{tabular}\end{table}

\bibliographystyle{elsarticle-harv}
\bibliography{references}